\documentclass{article}
\usepackage[utf8]{inputenc}
\usepackage{titlesec}
\usepackage{authblk}
\pdfoutput=1

\usepackage[sorting=none]{biblatex}
\usepackage{enumitem}
\usepackage{graphicx}
\usepackage[section]{placeins}
\usepackage{float}

\title{Automatic segmentation of prostate zones}
\author{Germonda Mooij, In\^{e}s Bagulho, Henkjan Huisman}
\affil{Radboud University Medical Centre Nijmegen}
\date{\today}

\begin{document}

\maketitle

\begin{abstract}
Convolutional networks have become state-of-the-art techniques for automatic medical image analysis, with the U-net architecture \cite{Unet} being the most popular at this moment. In this article we report the application of a 3D version of U-net \cite{Cicek} to the automatic segmentation of prostate peripheral and transition zones in 3D MRI images. Our results are slightly better than recent studies that used 2D U-net \cite{Padgett2016} and handcrafted feature \cite{Clark2017} approaches. 

In addition, we test ideas for improving the 3D U-net setup, by 1) letting the network segment surrounding tissues, making use of the fixed anatomy, and 2) adjusting the network architecture to reflect the anisotropy in the dimensions of the MRI image volumes. While the latter adjustment gave a marginal improvement, the former adjustment showed a significant deterioration of the network performance. We were able to explain this deterioration by inspecting feature map activations in all layers of the network. We show that to segment more tissues the network replaces feature maps that were dedicated to detecting prostate peripheral zones, by feature maps detecting the surrounding tissues.
\end{abstract}

\section{Introduction}

Prostate cancer is one of the major causes of cancer death for men in the Western world.
Multi-parameter MRI (mpMRI) is increasingly being used in the diagnosis of prostate cancer. Widespread implementation of mpMRI diagnosis of prostate cancer could avoid
unnecessary biopsies and enable population screening of prostate cancer, allowing early diagnosis of
the disease. However, widespread implementation is impeded because there is a lack of standardisation and because it requires substantial expertise in reading MRI scans. Automation of the detection of prostate tumor lesions
in mpMRI scans would be instrumental to overcome these impediments.

\subsection{Background}
Automatic segmentation of biomedical images has taken flight in recent years thanks to the development
of deep learning techniques. So-called convolutional neural networks (CNNs) significantly
outperform classical techniques like hand-made feature mapping. The current state of the art is the U-net
architecture \cite{Unet}, which has been successfully applied to segmentation problems for various organs
and imaging modalities. The power of U-net is that it captures both contextual and local information
to segment both full organs and the details of the organ’s borders.
The method has been successfully applied to entire prostate segmentation on mpMRI volumes \cite{Milletari}, and in this project this has been extended to segment the peripheral zone (PZ) and transition zones (TZ) of the prostate separately. This is challenging because the border between zones is harder to identify and enhancements of the method may be required. The medical significance of segmenting the zones is that the guidelines for mpMRI diagnosis of prostate cancer \cite{PIRADS} are different depending on which zone of the prostate the tumor/lesion is located. Therefore any technique for automating the detection of prostate lesions should include a zonal segmentation step.

\subsection{Related work}
Recent reviews (\cite{Shen2017}, \cite{PROMISE12}) have highlighted that deep learning, and convolutional networks in particular \cite{cs231}, has been applied to a wide range of medical image analysis tasks (segmentation, classification, detection,registration, image reconstruction, enhancement, etc.) across a wide range of
anatomical sites (brain, heart, lung, abdomen, breast, prostate, musculature, etc.).

Full prostate segmentation using a V-net architecture has been reported by Milletari \cite{Milletari}, achieving a dice score of 0.87 on the PROMISE2012 challenge dataset \cite{PROMISE12}. A 3D U-net architecture \cite{Cicek} was used for the prostate segmentation of another set of MRI scans, and an improvement of dice score from 90.1 to 92.7 was obtained using both scans of axial and sagittal orientations instead of only scans of the axial orientation \cite{Meyer2018}.

Segmentation of the prostate PZ and TZ separately has been reported using both ATLAS methods \cite{Padgett2016} (dice score 0.83 for prostate and 0.57 for PZ) and using a 2-stage 2D U-net method \cite{Clark2017} (dice score $0.92 \cdot 0.89=0.82$ for prostate, and $0.92 \cdot 0.84=0.77$ for TZ).

\subsection{Optimising deep learning for medical images} \label{sec12}
Here we test ideas for improving the 3D U-net setup for the task of segmenting the prostate PZ and TZ in mpMRI prostate scans.

We hypothesize that certain characteristics of medical images can be exploited to improve deep learning strategies. One important characteristic is that medical images always have the same topology, given by the anatomy of the organ of interest and the surrounding tissues. We speculate that the network can improve segmentation predictions by taking these surroundings into account. This can be achieved by providing the network with a large input image volume that includes the surroundings and by annotating surrounding tissues.

Another characteristic is the dimensions of the 3D volumes. In the case of MRI scans, image volumes are high resolution in one plane (e.g. 0.5x05mm in the axial plane), while in the third direction the slices are much thicker (e.g. 3.6 mm).
We suggest that segmentation performance will improve if the anisotropy of the input dimensions is reflected in an anisotropic 3D U-net. 

\section{Methods}

\subsection{Data volumes and manual segmentations}

\begin{figure}[h!]
    \centering
    \includegraphics[width=0.4\textwidth,trim={0cm 0cm 0cm 0cm},clip]{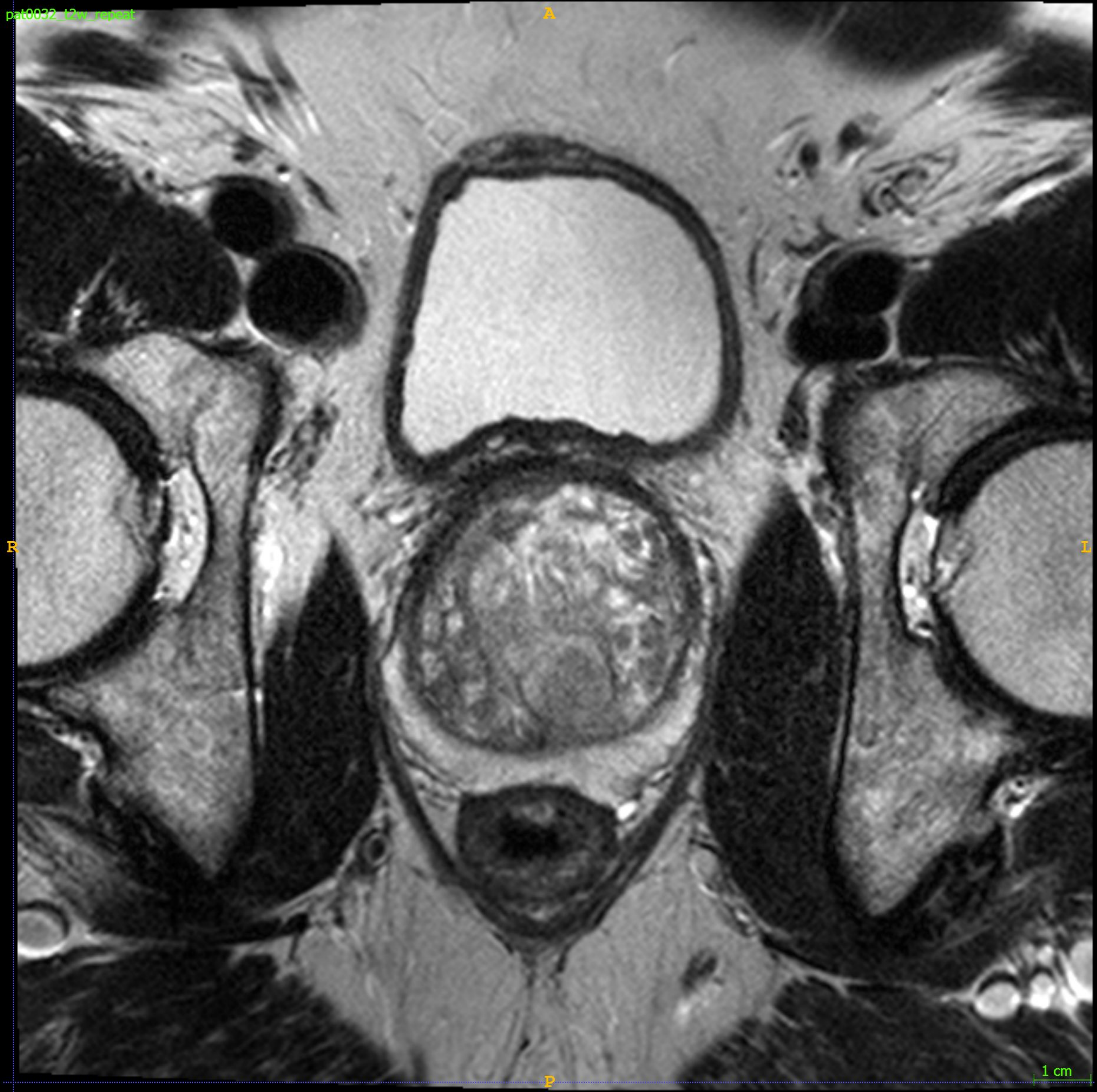}
    \includegraphics[width=0.4\textwidth,trim={0cm 0cm 0cm 0cm},clip]{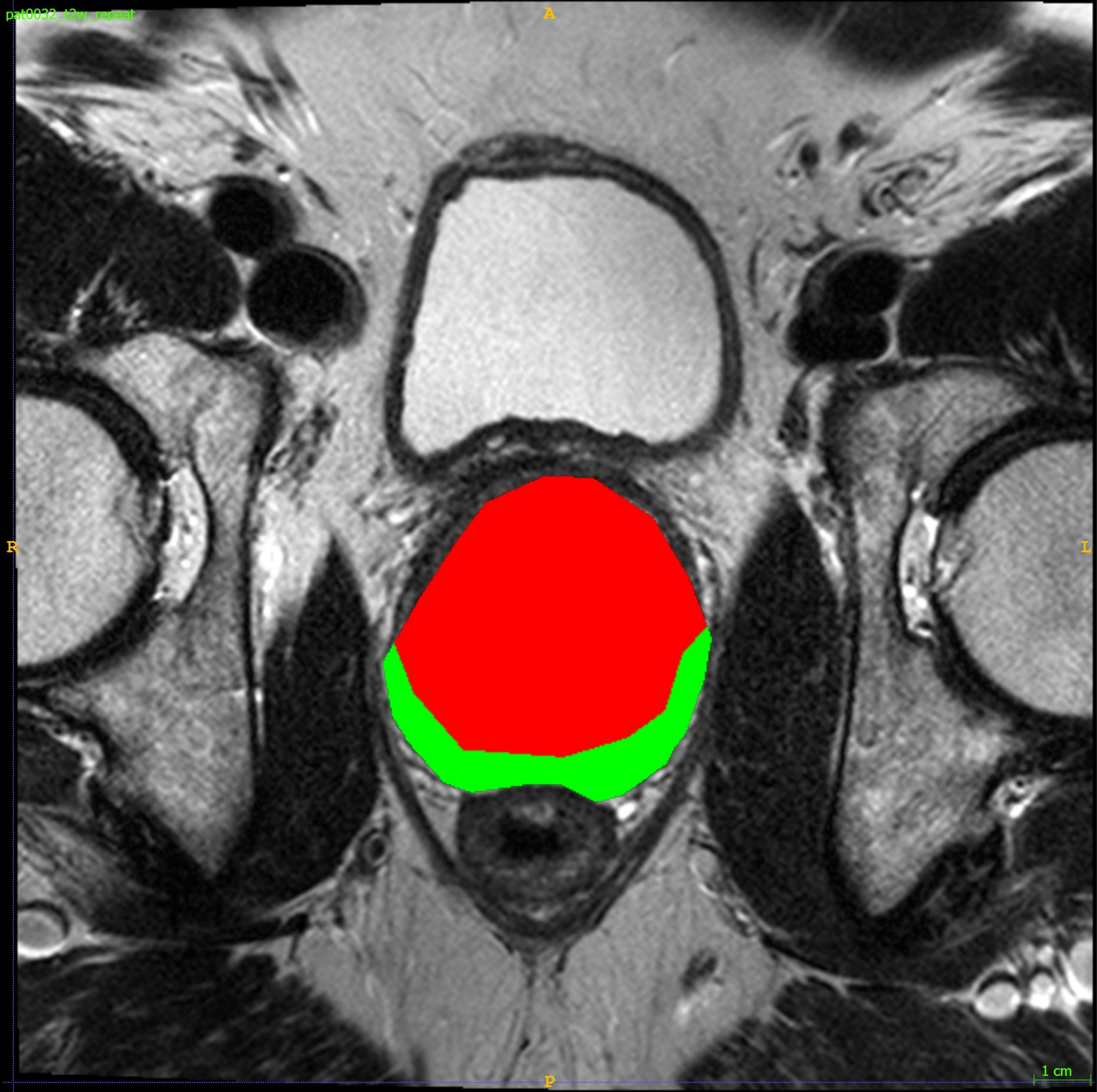}\\
\caption{Original image (left) and manual segmentation (right) of one of the test volumes. Colours are red for TZ and green for PZ.}
\label{fig:2label_manual}
\end{figure}

The data set used in this project consists of fifty-three 3D T2-weighted MRI volumes of the prostate and surrounding tissues from a large prostate mpMRI Scientific Archive \cite{detection2016}. 
In all volumes, the prostates were annotated by hand, distinguishing between the prostate PZ and TZ as shown in Figure \ref{fig:2label_manual}. While manual annotation was done in the axial view, a sagittal view was consulted for especially the apex and the base which can be hard to identify in the axial view. \\

\begin{figure}[h!]
    \centering
    \includegraphics[width=\textwidth,trim={0cm 6cm 0cm 0cm},clip]{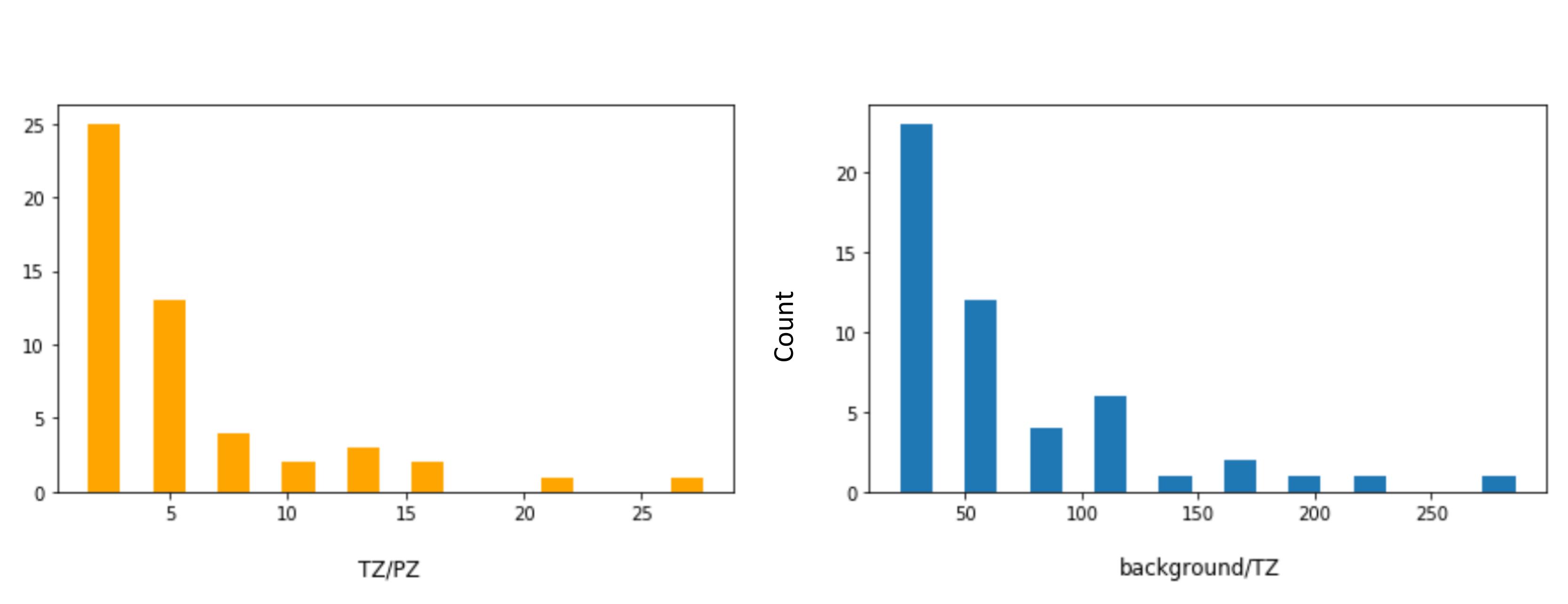}\\
\caption{Distribution of volume balances for TZ versus PZ (left) and for background versus TZ (right). This shows that the data set is highly imbalanced, and that the PZ is underrepresented.}
\label{fig:PZ_TZ_balances_51vols}
\end{figure}

The original image resolution of 0.5x0.5x3.6 mm and 0.3x0.3x3.6 mm was resampled to 1x1x3.6mm to fit the full image into GPU memory. 
The data set is imbalanced, with the PZ under-represented as shown in Figure \ref{fig:PZ_TZ_balances_51vols}. 

\begin{figure}[h!]
    \centering
    \includegraphics[width=0.4\textwidth,trim={0cm 0cm 0cm 0cm},clip]{pat0032_img.pdf}
    \includegraphics[width=0.4\textwidth,trim={0cm 0cm 0cm 0cm},clip]{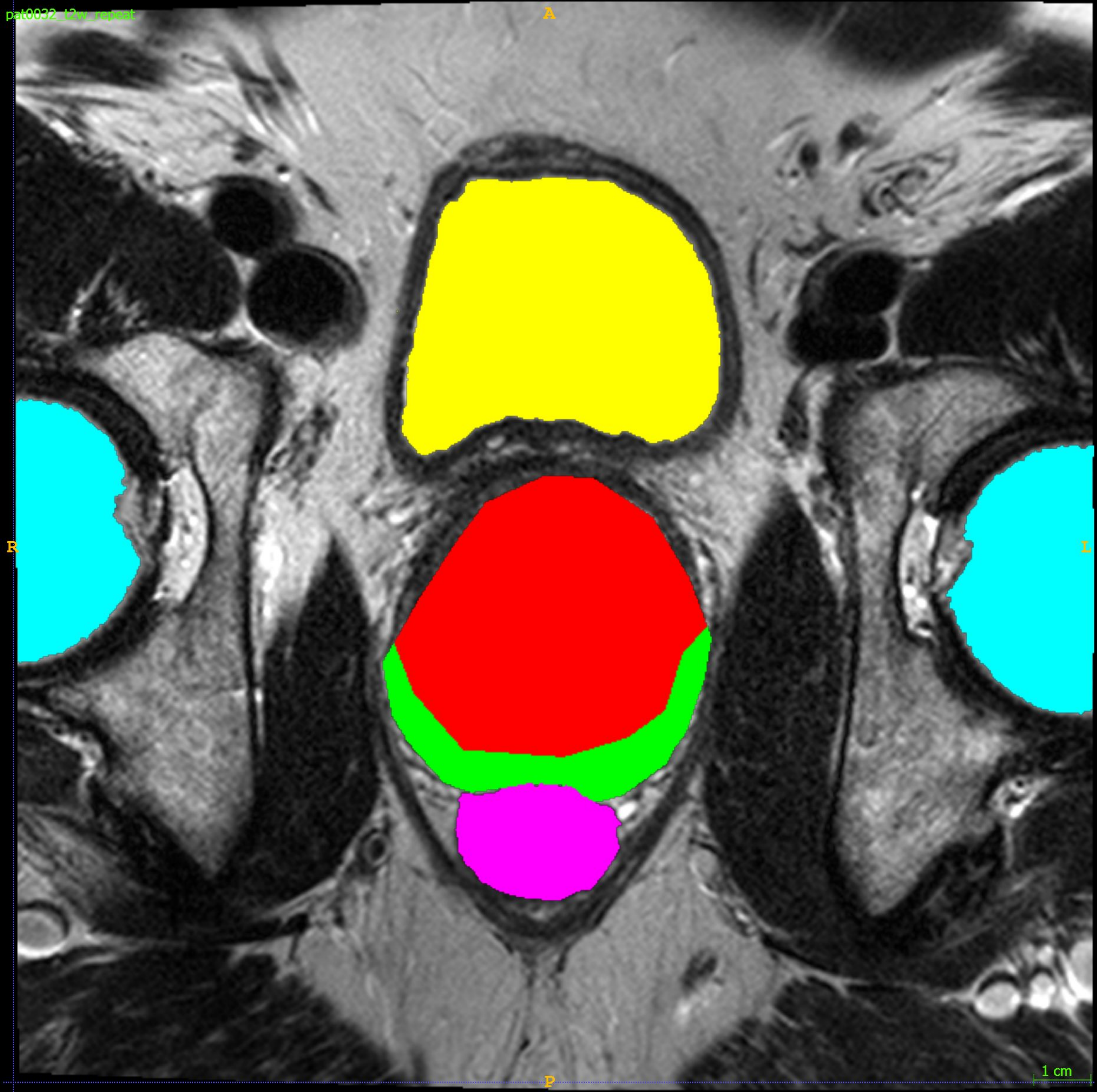}\\
\caption{Original image (left) and manual segmentation (right) of one of the test volumes. Colours are red for TZ, green for PZ, yellow for bladder, magenta for rectum, cyan for femur bones.}
\label{fig:5label_manual}
\end{figure}

In addition the bladder, rectum and femur bones were annotated as shown in Figure \ref{fig:5label_manual}.\\

\subsection{Network architecture}
\begin{figure}[h!]
    \centering
    \includegraphics[width=\textwidth,trim={3cm 1.5cm 2cm 1.5cm},clip]{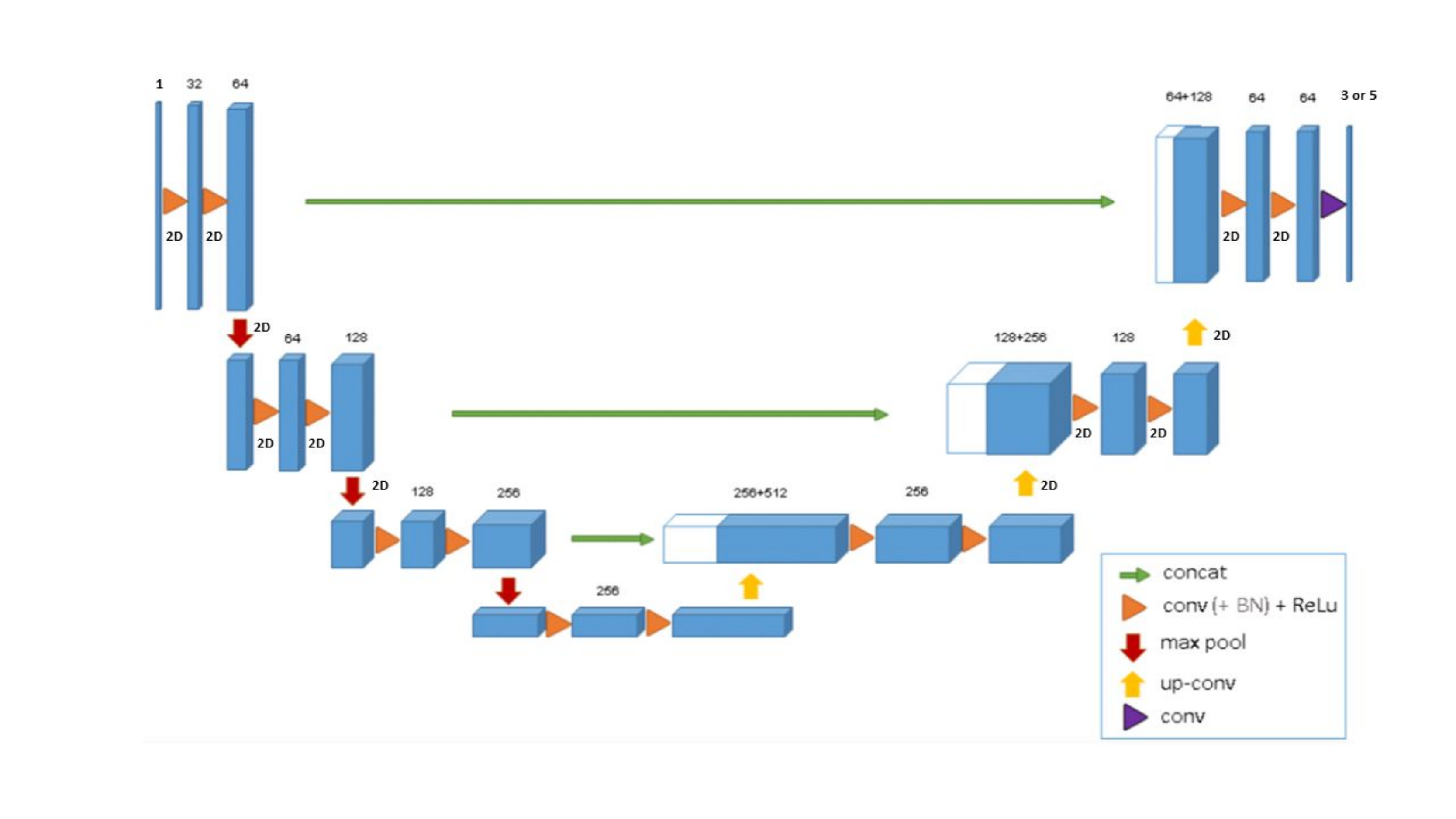}\\
\caption{The aniso-3DUNET architecture that is used in this project. This figure is taken from {\c{C}}i{\c{c}}ek et al.\cite{Cicek} with changes to 2D convolutions and 2D maxpoolings annotated.}
\label{fig:anisotropic3Dunet}
\end{figure}

Our anisotropic network, aniso-3DUNET, follows the 3D U-net architecture by {\c{C}}i{\c{c}}ek \cite{Cicek}, with the difference that it starts and ends with two layers of two 2D convolutions and one 2D maxpooling each (see Figure \ref{fig:anisotropic3Dunet}). The final step is a softmax either to 3 labels (background, TZ, and PZ) or 6 labels (background, TZ, PZ, bladder, rectum and femur bones), preceded by two steps for each voxel that map 64 to 64 and then 64 to 3 or 5 features (depending on the number of tissues that are segmented)).

An alternative more isotropic model, iso-3DUNET, is evaluated, in which all the maxpoolings are 3D as in the original 3D U-net. In view of the low dimensionality in the vertical direction, every second convolution in each layer is a 2D convolution instead of a 3D convolution as in the original 3D U-net.  

\begin{figure}[h!]
    \centering
    \includegraphics[width=\textwidth,trim={3cm 1.5cm 2cm 1.5cm},clip]{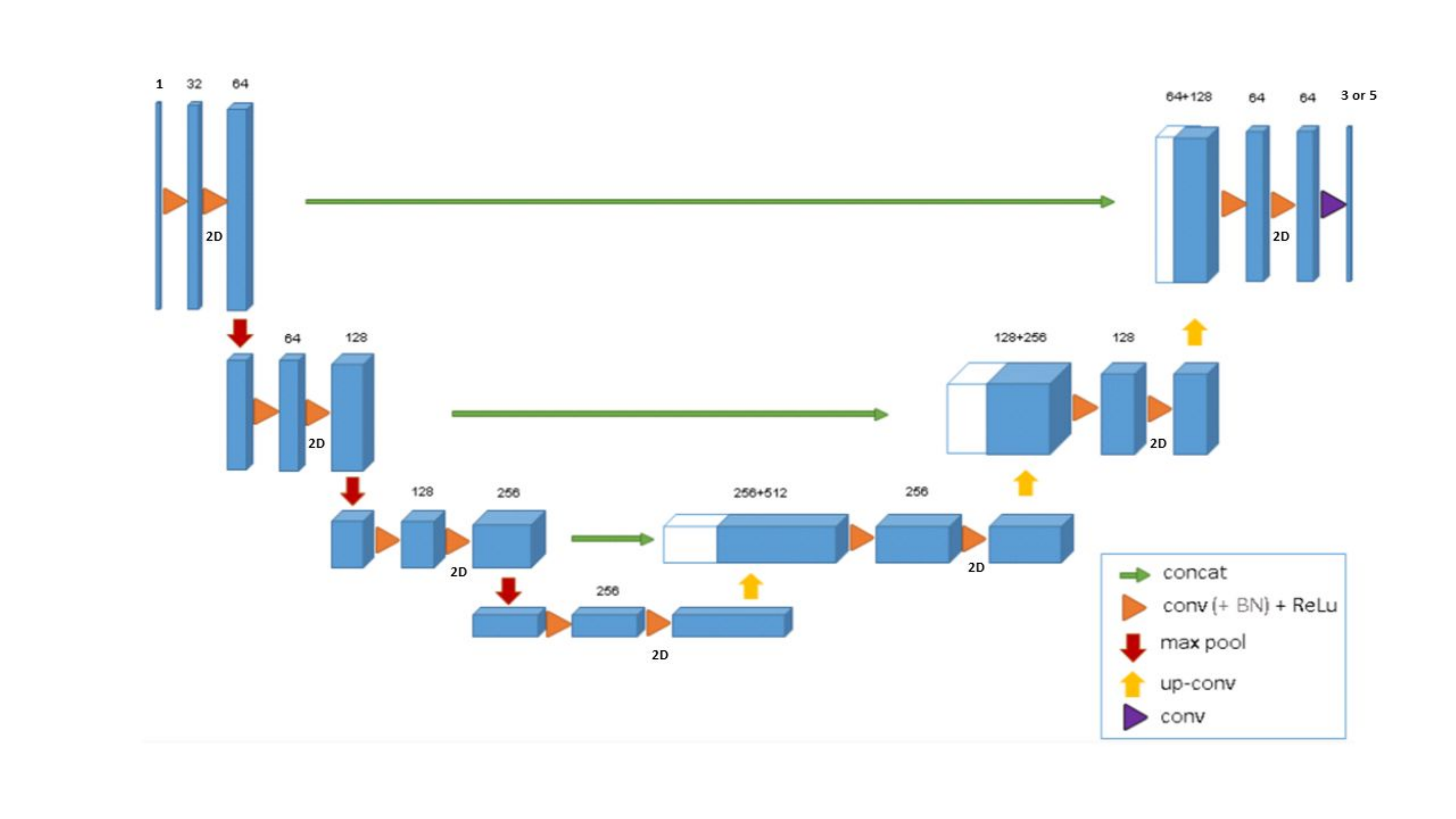}\\
\caption{The alternative iso-3DUNET architecture that is evaluated in this project. This figure is taken from {\c{C}}i{\c{c}}ek et al.\cite{Cicek} with changes to 2D convolutions annotated.}
\label{fig:isotropic3Dunet}
\end{figure}

\subsection{Training and testing}

During training, data augmentation was applied to the images in the form of small translations, rotations, isotropic expansions and contractions, elastic deformations and left-right flips.

Due to the low number of annotated volumes available for training, it was expected that validation scores vary depending on the validation split, so a 5-fold cross validation was carried out to obtain a range of validation scores. Training was aimed at minimizing a multi-label cross entropy loss, and focussed on the organs of interest by weight factors to counter the data imbalance shown in Figure \ref{fig:PZ_TZ_balances_51vols}. Loss contributions from each voxel were given a weight linked to its ground truth label: background = 1, TZ = 2, PZ = 6, and other organs = 1. These numbers  do  reflect  the  skew  in  the  imbalance  but  are  lower  than  the  actual imbalances, because excessive weighting results in over-prediction of the zones. After each cross-validation run, predictions were generated for the validation volumes.

After the 5-fold cross-validation runs, one more model training was carried out on all the training volumes and predictions were generated on a set of eight test volumes that had been kept separate.

The models were trained using the Keras framework with a learning rate of 0.00001, glorot uniform initialization, L2 kernel regularization, and an Adam optimizer. The number of epochs for each training run is 300, which is limited by run time: the 6 training runs required to evaluate a scenario together take 5 days on a single GPU.

\subsection{Experiments}
Two sets of experiments were run:
\begin{enumerate}
    \item 
    The aniso-3DUNET model shown in Figure \ref{fig:anisotropic3Dunet} was trained on the same input volumes, first to segment only the prostate PZ and TZ (2-label case) and then to segment the prostate PZ and TZ, the bladder, rectum and femur bones (6-label case).
    \item
    The other comparison we make is between two different models: the aniso-3DUNET as shown in Figure \ref{fig:anisotropic3Dunet}, and the iso-3DUNET as shown in Figure \ref{fig:isotropic3Dunet}. 
\end{enumerate}
The metric used for scoring predictions for each data input volume was dice score:
\begin{equation}
    dice = \frac{2 |P \cap G|}{|P| + |G|}
\end{equation}
where $P$ is the volume of predicted segmentation probabilities and $G$ is the volume of ground truth labels.

The training and validation scores are reported as a function of the number of epochs. The distribution and average of the dice scores is also reported for the predictions of each of the fifty-three input volumes after the validation runs or the test run (for whichever run an input volume was not part of the training set).

To find out how the network is learning, the layers of the model are visualized. There are several ways to do this \cite{Chollet2017}, and here feature maps  are plotted in the form of activations for the second convolution in each layer for a single test input volume. 

\section{Results}

Here dice scores are reported for 5-fold cross-validation training runs and a test run, for each of the scenario's that we are evaluating. The dice scores at the end of most runs are still increasing, and longer runs would achieve somewhat higher dice scores. The highest scores are obtained by segmenting only TZ and PZ with the aniso-3DUNET shown in Figure \ref{fig:anisotropic3Dunet}: dice scores of 0.85 for TZ and 0.60 for PZ. The PZ is segmented most reliably in the middle of the prostate. For a split of the prostates into a base (top 1/5), middle (mid 3/5) and apex (bottom 1/5) sections, average dice scores of respectively 0.46, 0.71 and 0.51 are obtained. 

\subsection{Segmenting surrounding organs}

\begin{figure}[h!]
    \centering
    \includegraphics[width=\textwidth,trim={4cm 0.5cm 3cm 0cm},clip]{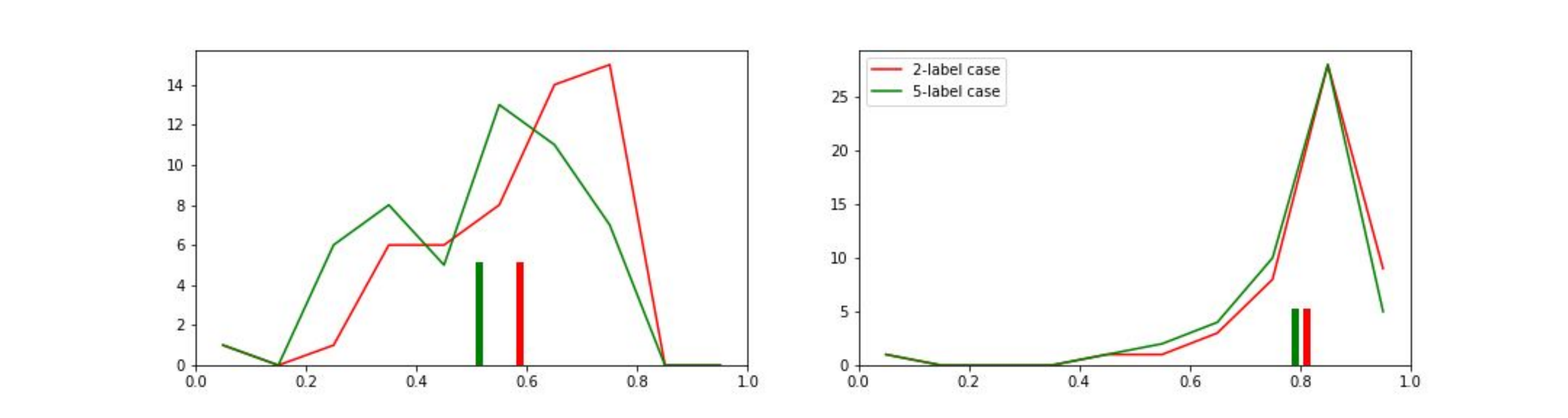}\\
    dice score \hspace{5cm} dice score
\caption{Distribution of dice scores for PZ (left) and TZ (right), comparing the 2-label case (red), with the 6-label case (green).}
\label{fig:hist_PZ_TZ_run_2lbl_5lbl}
\end{figure}

The distribution and average dice score for all fifty-three input volumes are compared in Figure \ref{fig:hist_PZ_TZ_run_2lbl_5lbl}. Clearly the results contradict our hypothesis in section \ref{sec12}, that segmenting more organs would benefit the prostate segmentation. The case where the prostate zones plus surrounding organs are segmented predicts the PZ significantly worse than the case where only the prostate zones are segmented. There is virtually no difference in predicting the TZ.

\begin{figure}[h!]
    \centering
    \includegraphics[width=\textwidth,trim={4cm 1.5cm 0cm 0cm},clip]{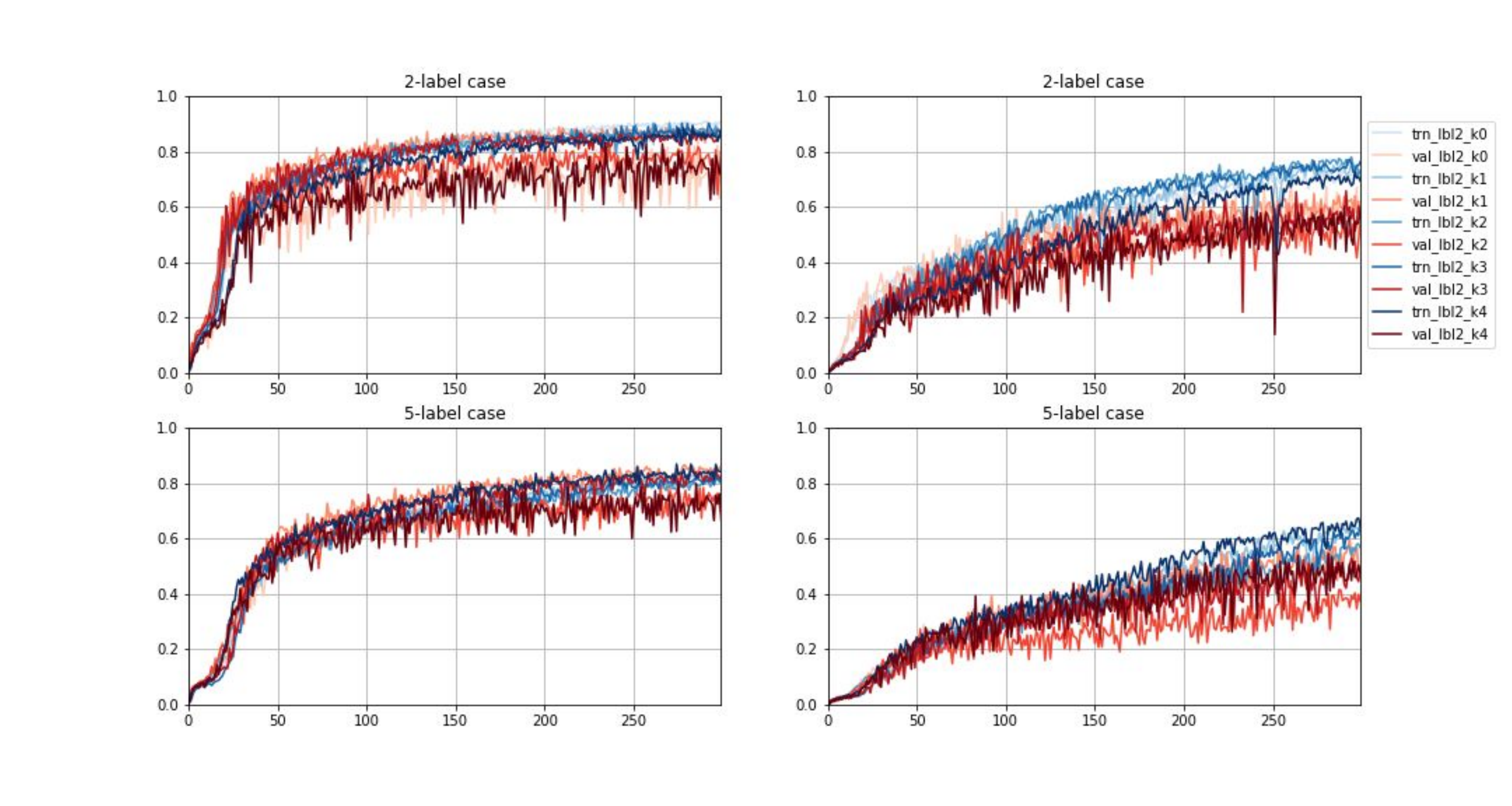}\\
    epochs \hspace{5cm} epochs
\caption{Dice scores achieved throughout the training and 5-fold cross-validation runs. To the left are TZ scores, to the right PZ zone scores. Blue curves are training (trn) and red curves are validation (val) scores, and the colour tone is varied for each of the 5 cross-validation runs (k0-k4).
The graphs at the top are for the 2-label case, and at the bottom for the 6-label case.
The dice scores at the end of most runs are still increasing, and longer runs would achieve somewhat higher dice scores.}
\label{fig:cv_2lbl_5lbl}
\end{figure}

The same can be seen in Figure \ref{fig:cv_2lbl_5lbl}, which shows that segmenting more organs also slows down the training and increases the range of scores in the cross-validation runs.  

\subsection{Anisotropic 3D U-net (aniso-3DUNET).}

\begin{figure}[h!]
    \centering
    \includegraphics[width=\textwidth,trim={4cm 0.5cm 3cm 0cm},clip]{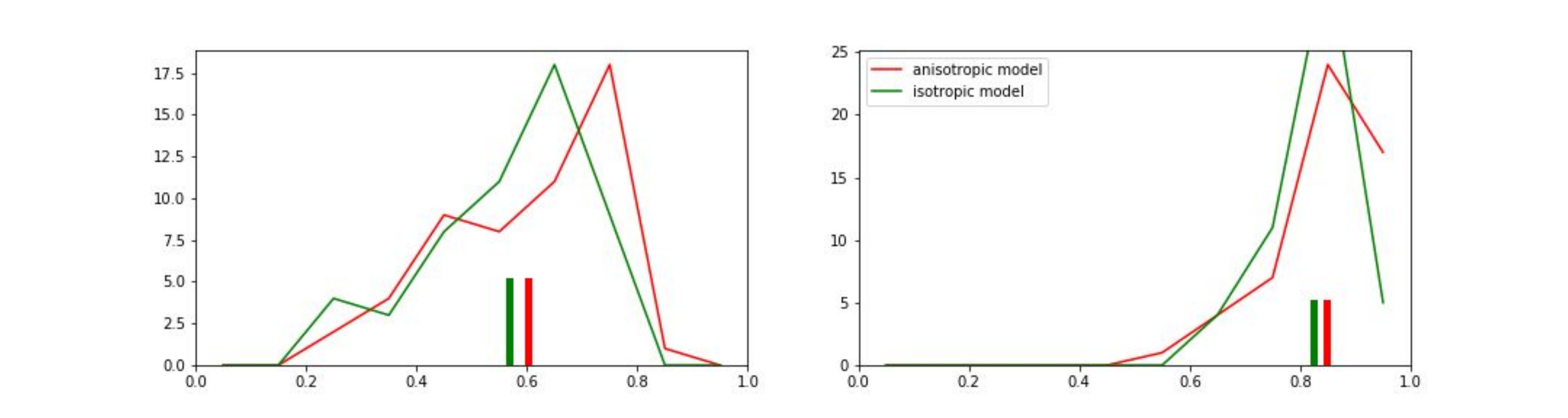}\\
    dice score \hspace{5cm} dice score
\caption{Distribution of dice scores for PZ (left) and TZ (right), comparing the aniso-3DUNET (red), with the iso-3DUNET (green).}
\label{fig:hist_PZ_TZ_run_iso_aniso}
\end{figure}

\begin{figure}[h!]
    \centering
    \includegraphics[width=\textwidth,trim={4cm 1.5cm 0cm 0cm},clip]{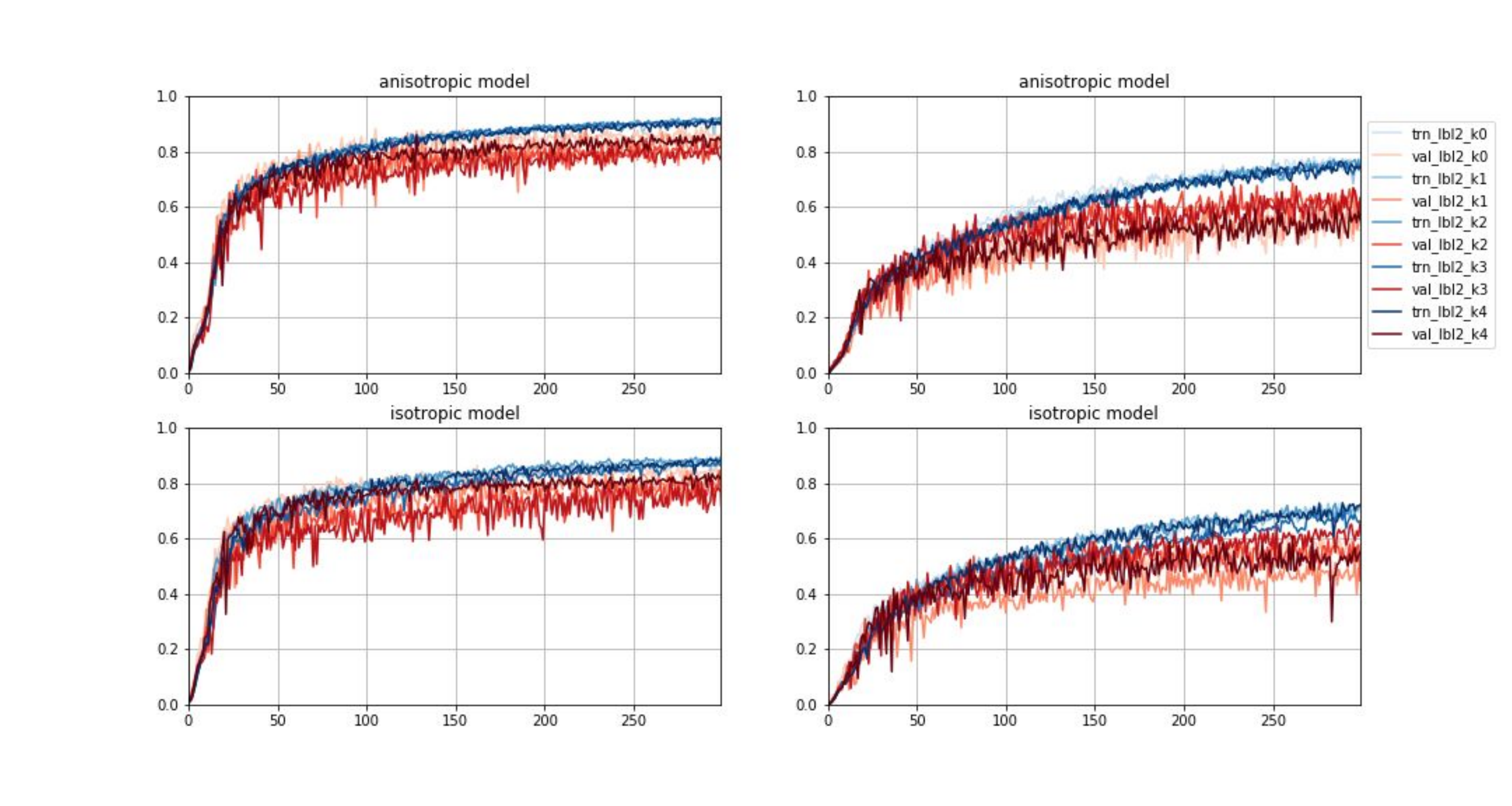}\\
    epochs \hspace{5cm} epochs
\caption{Dice scores achieved throughout the training and 5-fold cross-validation runs. To the left are transition zone scores, to the right peripheral zone scores. Blue curves are training (trn) and red curves are validation (val) scores, and the colour tone is varied for each of the 5 cross-validation runs (k0-k4).
The graphs at the top are for the aniso-3DUNET, and at the bottom for the iso-3DUNET.}
\label{fig:cv_iso_aniso}
\end{figure}

Figure \ref{fig:hist_PZ_TZ_run_iso_aniso} shows that the aniso-3DUNET achieves a higher dice score (dice=0.60) than the iso-3DUNET (dice=0.57). The 0.03 increase in dice score is marginal though, as it is of the same order of magnitude as the standard error in Figure \ref{fig:hist_PZ_TZ_run_iso_aniso} which is $\sigma / \sqrt(N=51) \sim 0.14 / 7 \sim 0.02$. Figure \ref{fig:cv_iso_aniso} shows the variation between the cross-validation runs, which like in Figure \ref{fig:hist_PZ_TZ_run_iso_aniso} shows a minor improvement. 

\subsection{Feature map analysis.}

To understand how the networks learn and to further understand differences in performance, we plot the activations that are output by the second convolution in each layer. A slice map through the middle of the prostate is plotted for each of the features per layer, and the complete sets are shown in the Appendix. In Figure \ref{fig:feature_maps_2lbl} three of the features for each layer are shown for the best performing model (aniso-3DUNET with 2-label segmentation), which are selected because they provide some insights into how the network performs the segmentation. 
The input image is fed into the network at the top left layer. The information flow follows the red and green arrows, which are the same as the ones in the network graph shown in Figure \ref{fig:anisotropic3Dunet}. The flow ends at the last layer at the top right, which feeds into the softmax layer for final segmentation.

Based on Figure \ref{fig:feature_maps_2lbl} we try to deduce per layer what these features are coding for.

\begin{figure}[h!]
    \centering
    \begin{tabular}{c c c c}
    layer 1 &&& layer 7
    \end{tabular}\\
    \includegraphics[width=0.15\textwidth,trim={0cm 0cm 0cm 0cm},clip]{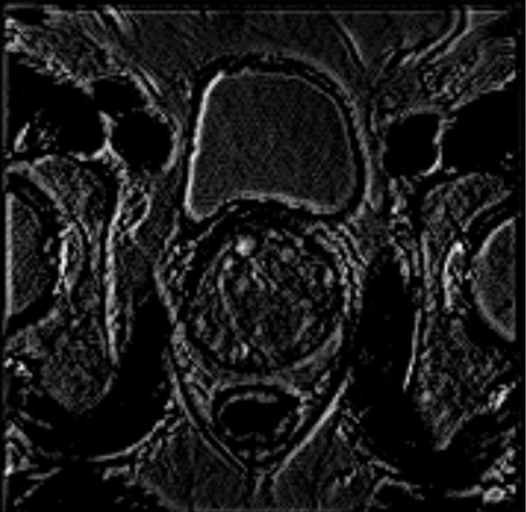}
    \includegraphics[width=0.15\textwidth,trim={0cm 0cm 0cm 0cm},clip]{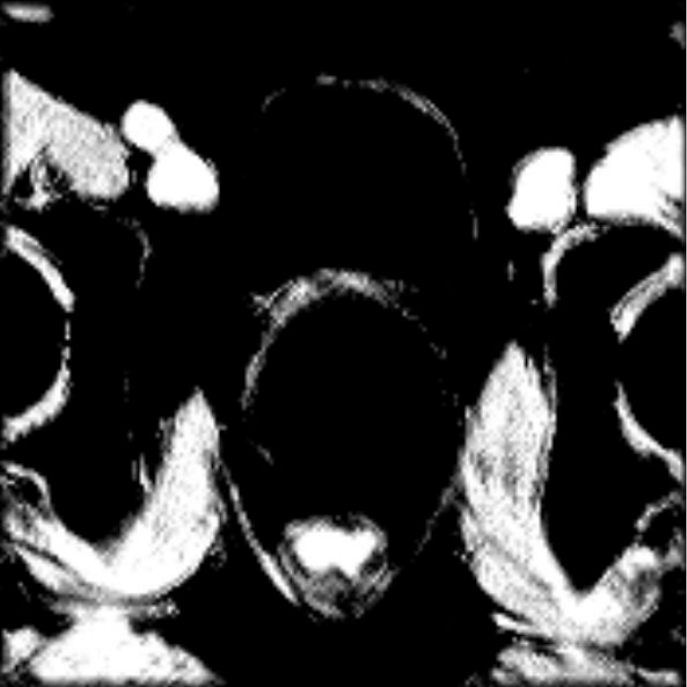}
    \includegraphics[width=0.15\textwidth,trim={0cm 0cm 0cm 0cm},clip]{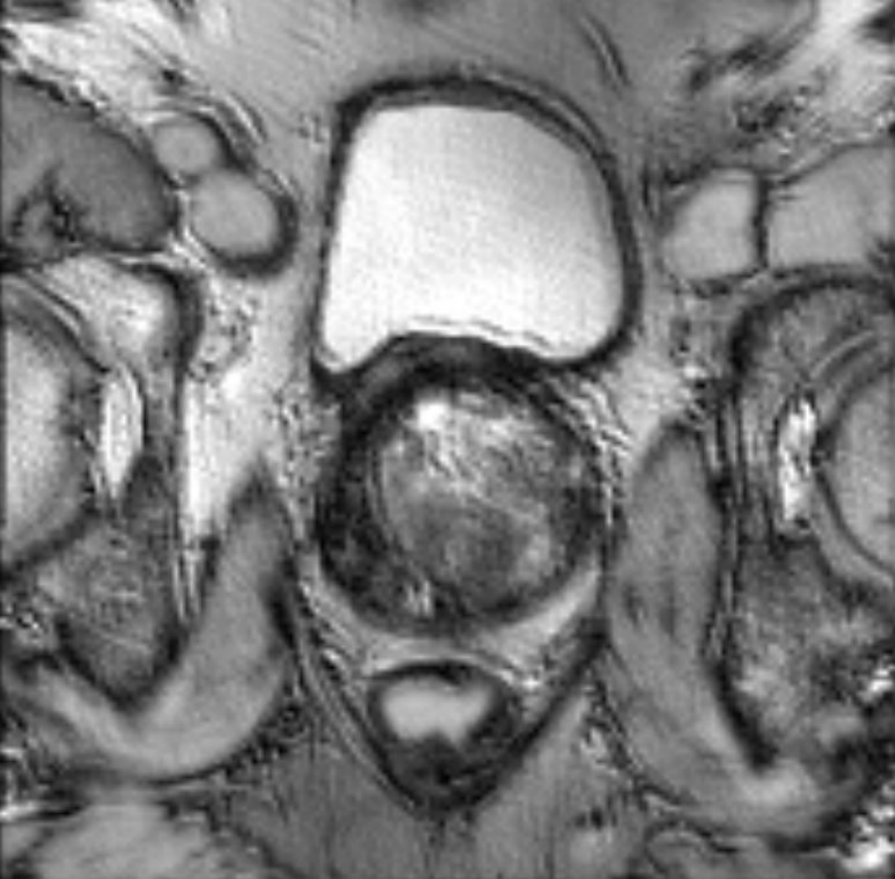}
    \includegraphics[width=0.03\textwidth,trim={0cm 0.2cm 0cm 0cm}, clip]{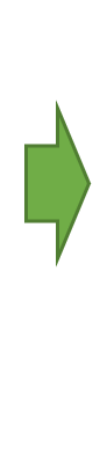}
    \includegraphics[width=0.15\textwidth,trim={0cm 0cm 0cm 0cm},clip]{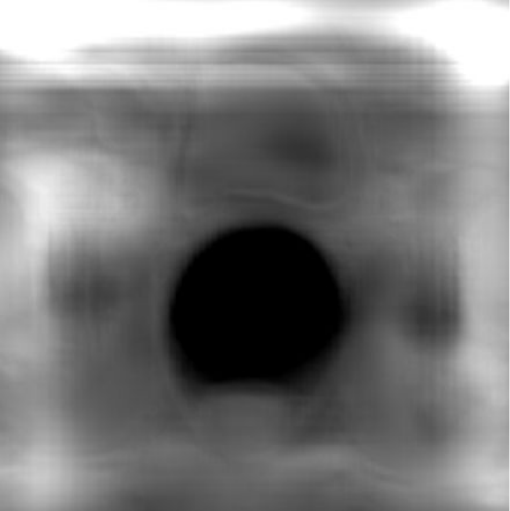}
    \includegraphics[width=0.15\textwidth,trim={0cm 0cm 0cm 0cm},clip]{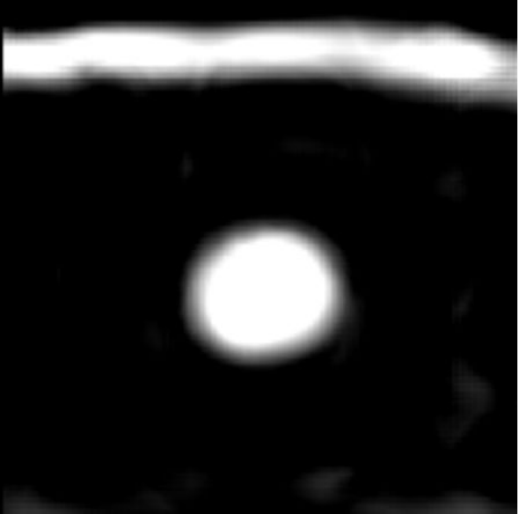}
    \includegraphics[width=0.15\textwidth,trim={0cm 0cm 0cm 0cm},clip]{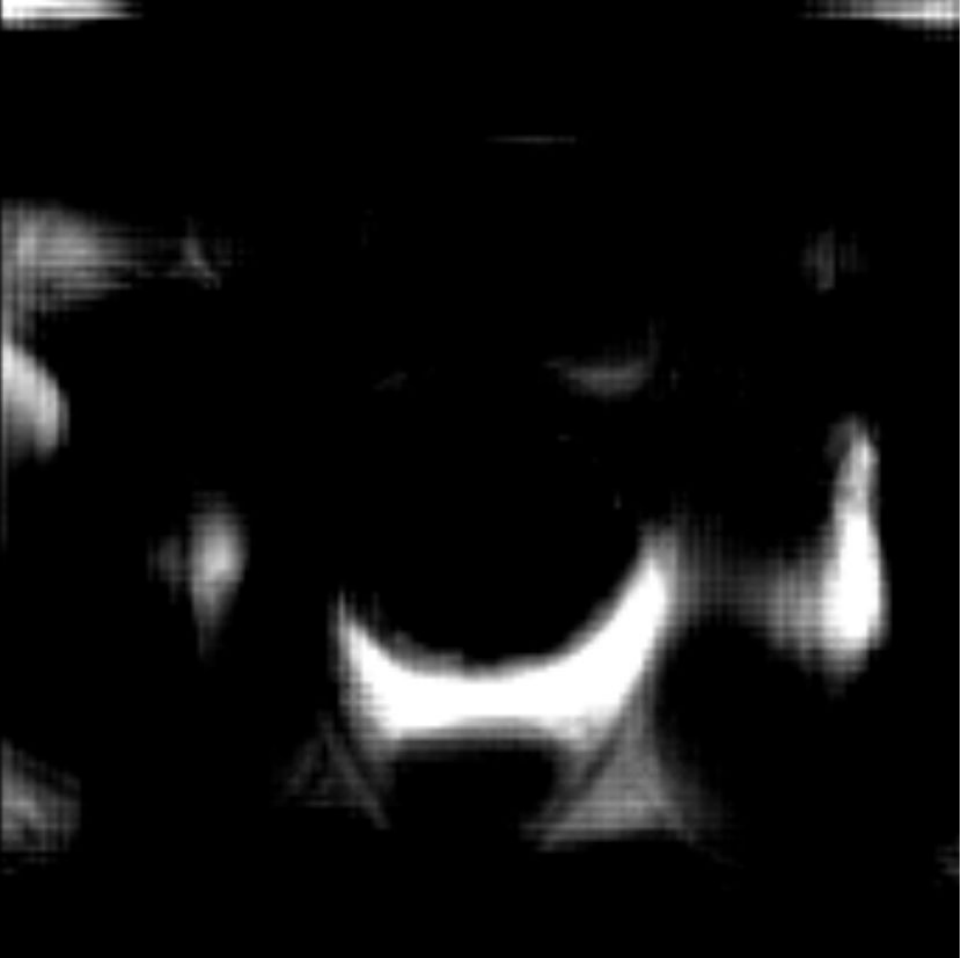}\\
    \includegraphics[width=0.3\textwidth,trim={0cm 0cm 5cm 0cm}, clip]{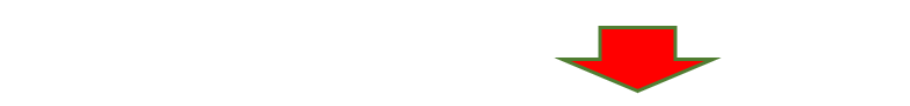}
    \begin{tabular}{c c c c}
    layer 2 &&& layer 6
    \end{tabular}
    \includegraphics[width=0.3\textwidth,trim={3cm 0cm 0cm 0cm}, clip]{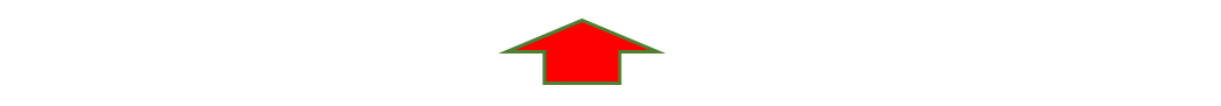}\\
    \includegraphics[width=0.15\textwidth,trim={0cm 0cm 0cm 0cm},clip]{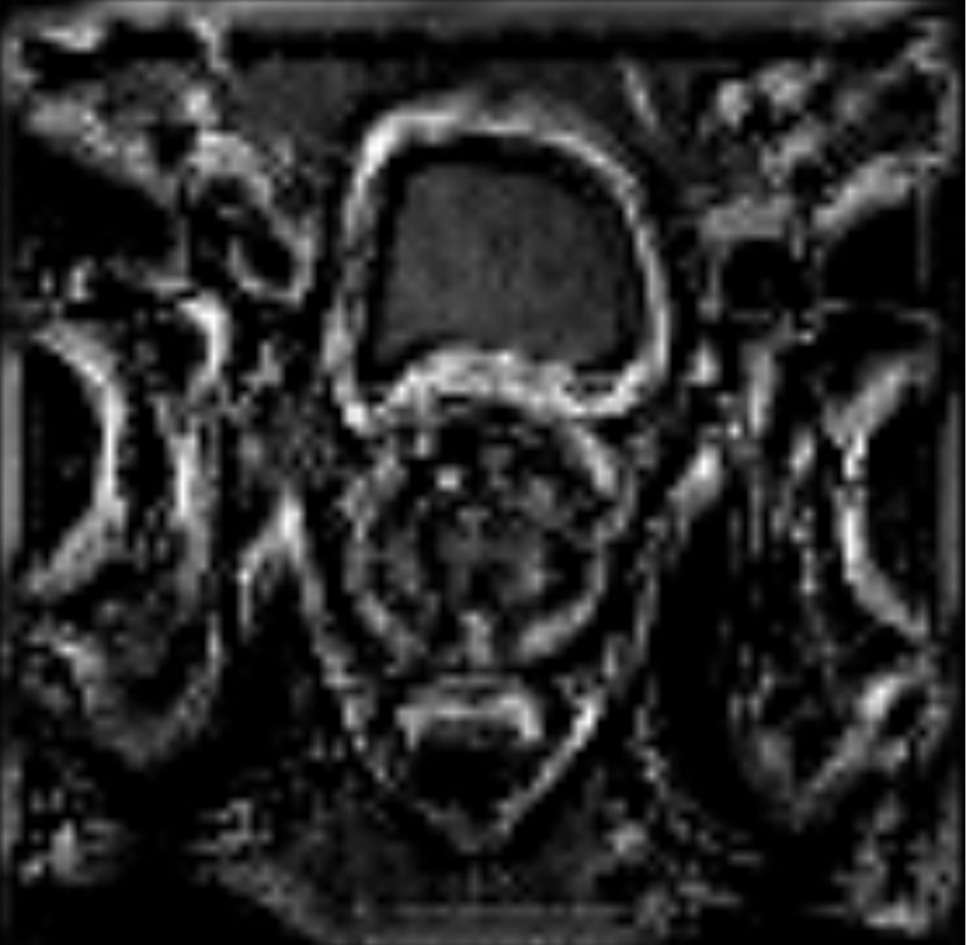}
    \includegraphics[width=0.15\textwidth,trim={0cm 0cm 0cm 0cm},clip]{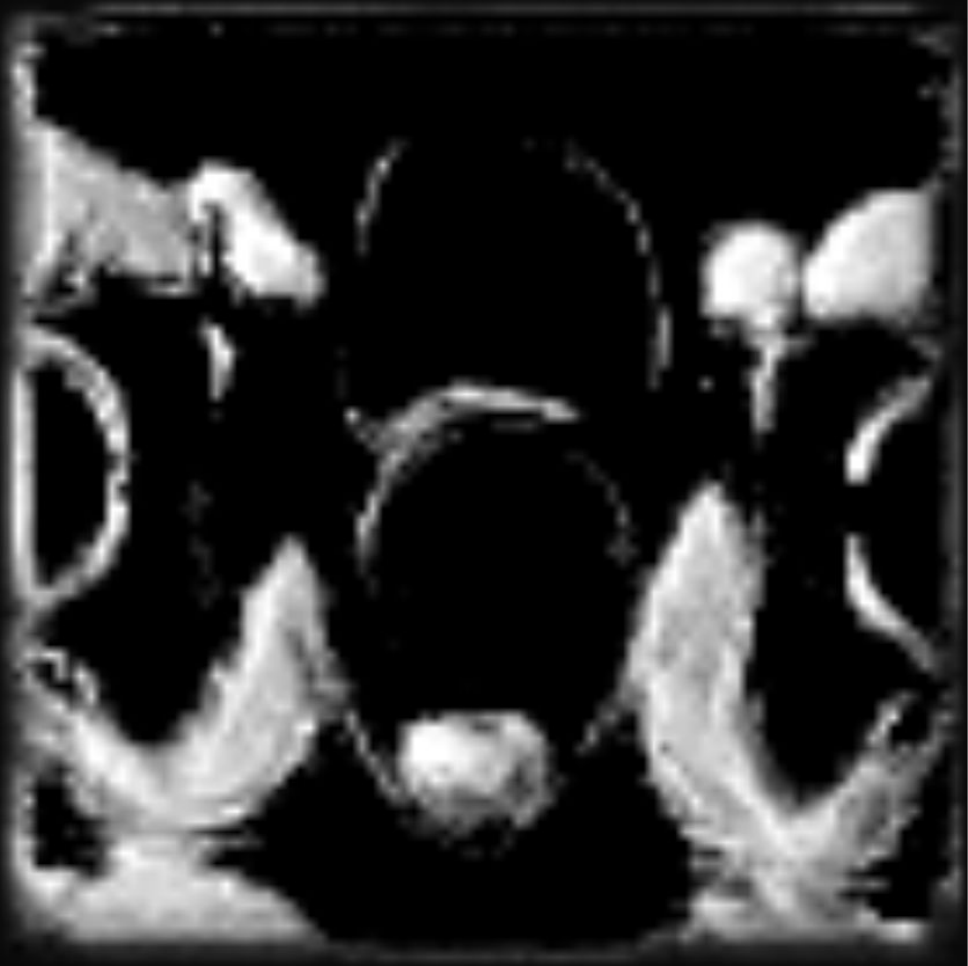}
    \includegraphics[width=0.15\textwidth,trim={0cm 0cm 0cm 0cm},clip]{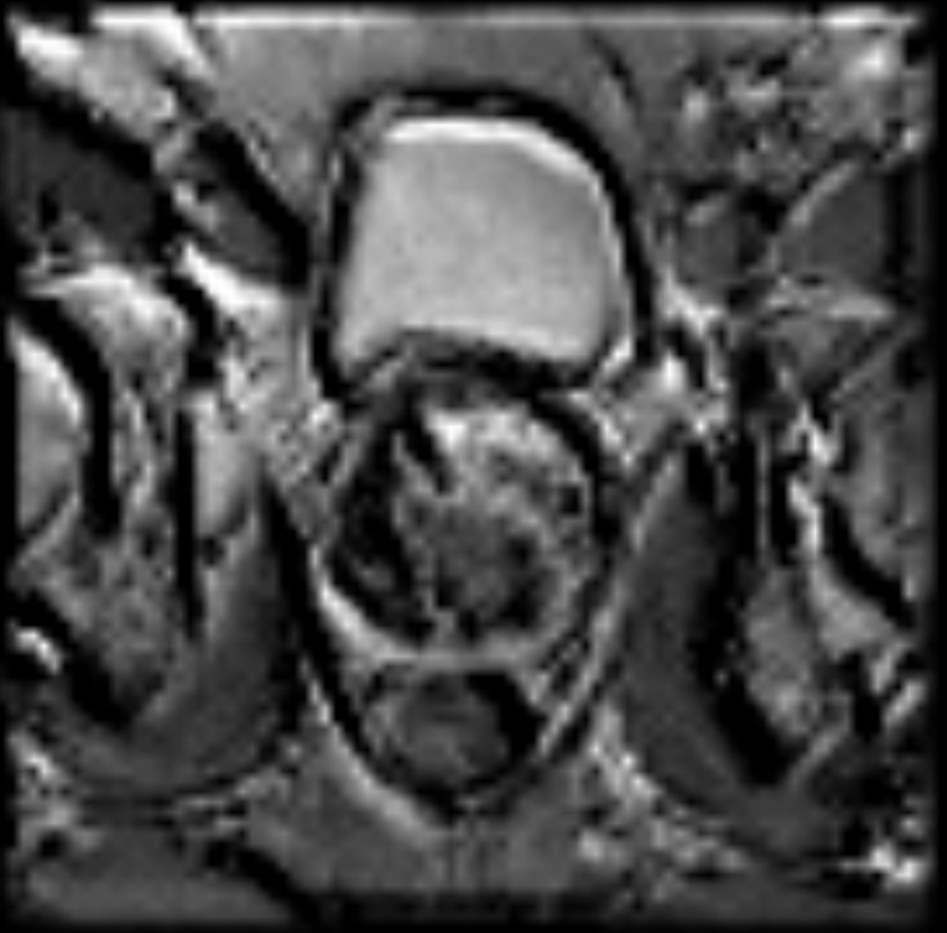}
    \includegraphics[width=0.03\textwidth,trim={0cm 0.2cm 0cm 0cm}, clip]{green_arrow_right.pdf}
    \includegraphics[width=0.15\textwidth,trim={0cm 0cm 0cm 0cm},clip]{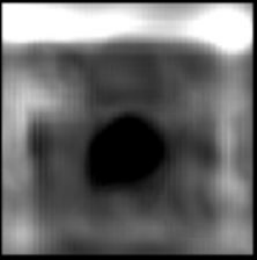}
    \includegraphics[width=0.15\textwidth,trim={0.1cm 0cm 0.1cm 0cm},clip]{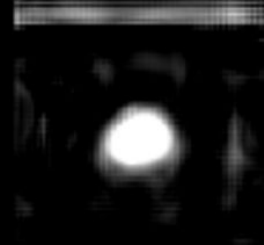}
    \includegraphics[width=0.15\textwidth,trim={0cm 0cm 0cm 0cm},clip]{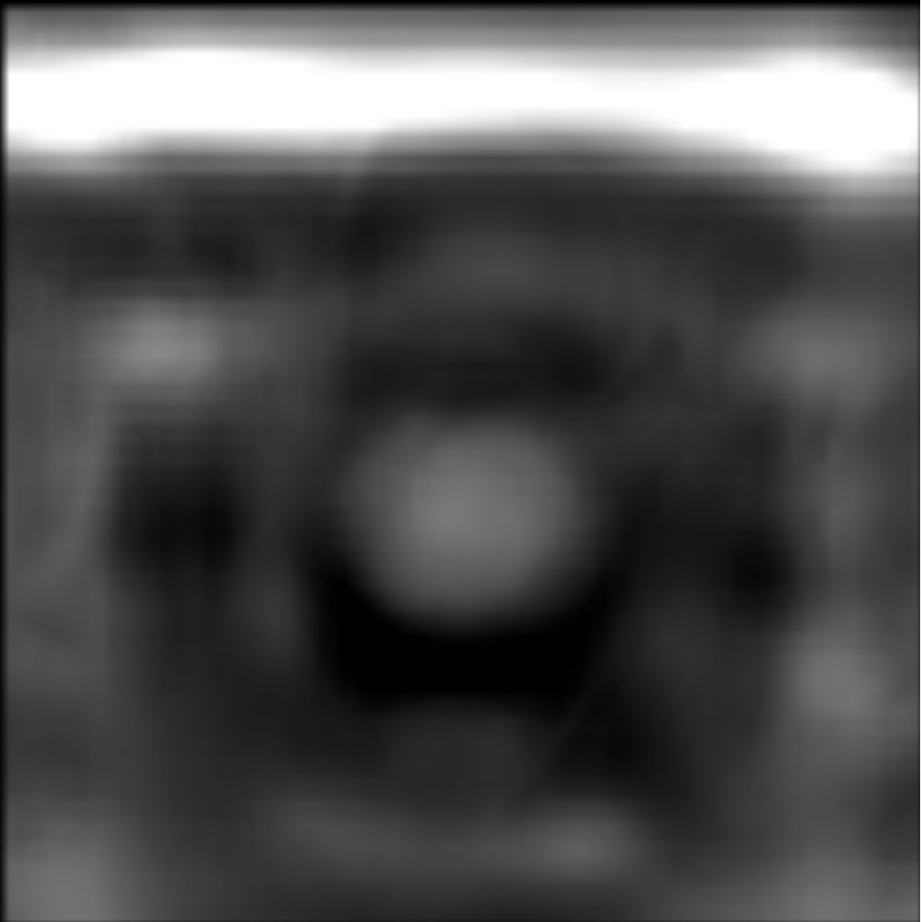}\\
    \includegraphics[width=0.3\textwidth,trim={0cm 0cm 5cm 0cm}, clip]{red_arrow_down.pdf}
    \begin{tabular}{c c c c}
    layer 3 &&& layer 5
    \end{tabular}
    \includegraphics[width=0.3\textwidth,trim={3cm 0cm 0cm 0cm}, clip]{red_arrow_up.pdf}\\
    \includegraphics[width=0.15\textwidth,trim={0cm 0cm 0cm 0cm},clip]{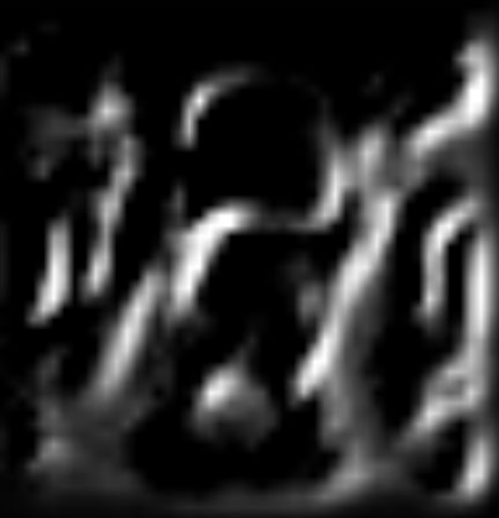}
    \includegraphics[width=0.15\textwidth,trim={0.5cm 0.5cm 0.5cm 0.5cm}, clip]{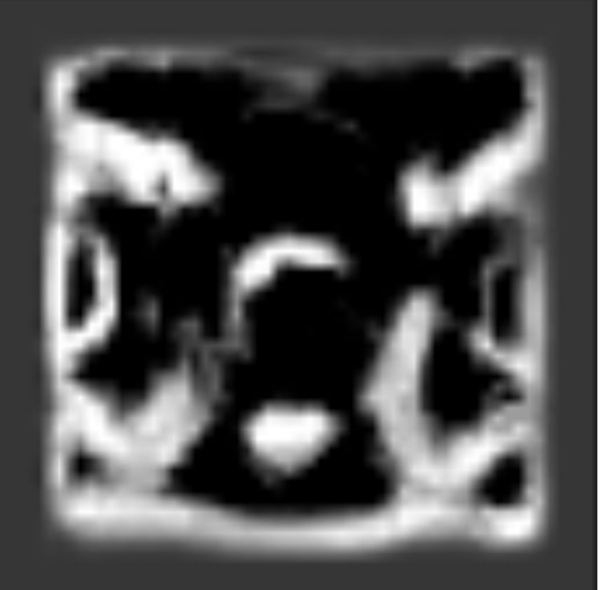}
    \includegraphics[width=0.15\textwidth,trim={0.5cm 0.5cm 0.5cm 0.5cm}, clip]{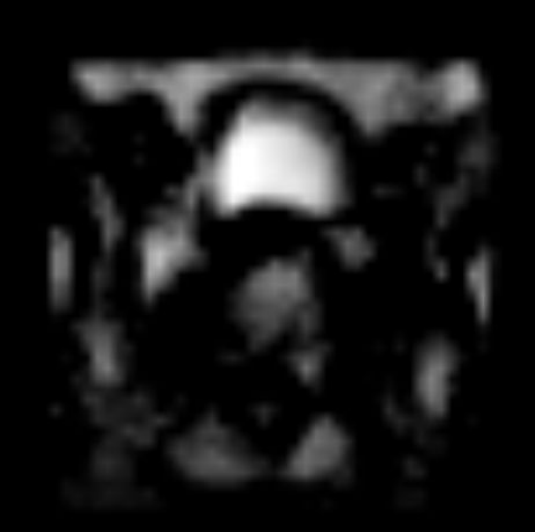}
    \includegraphics[width=0.03\textwidth,trim={0cm 0.2cm 0cm 0cm}, clip]{green_arrow_right.pdf}
    \includegraphics[width=0.15\textwidth,trim={0cm 0cm 0cm 0cm},clip]{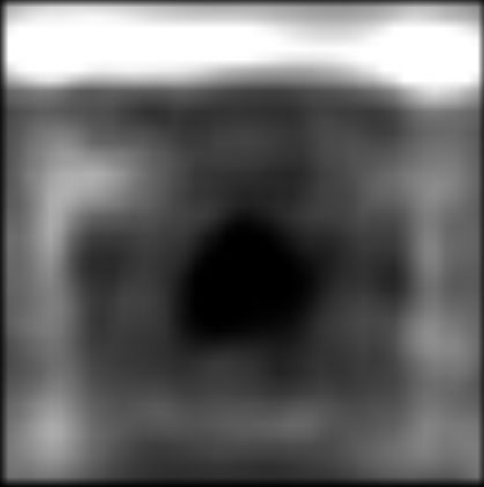}
    \includegraphics[width=0.15\textwidth,trim={0cm 0cm 0cm 0cm},clip]{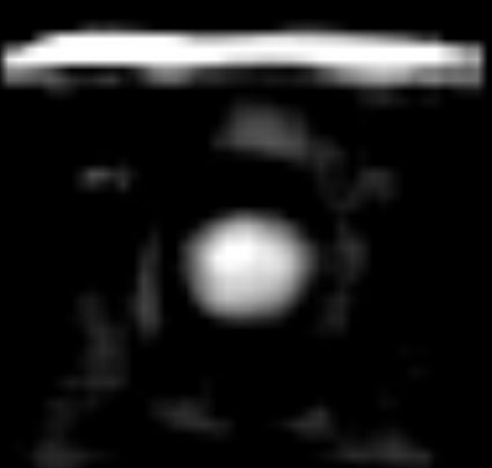}
    \includegraphics[width=0.15\textwidth,trim={0cm 0cm 0cm 0cm},clip]{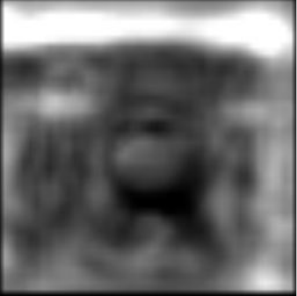}\\
    \includegraphics[width=0.3\textwidth,trim={00cm 0cm 6cm 0cm}, clip]{red_arrow_down.pdf}
        \begin{tabular}{c}
    layer 4
    \end{tabular}
    \includegraphics[width=0.3\textwidth,trim={4cm 0cm 0cm 0cm}, clip]{red_arrow_up.pdf}\\
    \includegraphics[width=0.15\textwidth,trim={0.1cm 0.1cm 0cm 0cm},clip]{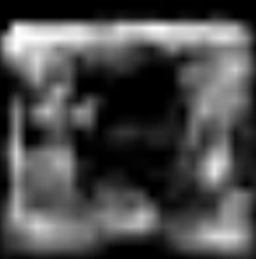}
    \includegraphics[width=0.15\textwidth,trim={0cm 0cm 0cm 0.1cm},clip]{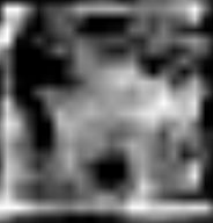}
    \includegraphics[width=0.15\textwidth,trim={0cm 0.1cm 0cm 0.1cm},clip]{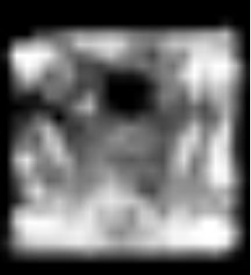}\\
\caption{Examples of three interesting features generated by the second convolution at each of the seven layers in the aniso-3DUnet segmenting 2 labels.}
\label{fig:feature_maps_2lbl}
\end{figure}

\begin{itemize}
\item[layer 1]
The three feature maps for the first model layer show that, as reported for other networks elsewhere \cite{Chollet2017}, this layer codes for local features like edges and thick black shapes at resp. the left and the middle feature maps. An interesting feature map is the one on the right, in which thick black shapes have been greyed out, but not thin black lines.  
\item[layer 2]
The second layer shows similar feature maps as the first layer, at the lower resolution of this layer.
\item[layer 3]
The feature maps in the third layer still show some similarity to the ones in the earlier layers, but are now more cartoon like. The feature map at the left is covering parts of the areas that are background in a white blur. The middle feature map is progressing the black shape feature maps in the previous layer, with the outline of the prostate shape thickened. The feature map at the right is showing both the bladder and the prostate TZ.
\item[layer 4]
This layer is the most downscaled layer, and the feature maps are very blurry white. Most maps look unstructured, but the three shown here do have some structure: the left feature map blurs the background areas white, the middle feature map blurs most of the image white except the rectum, and the feature map on the right blurs everything white except the centre of the bladder.
\item[layer 5]
The resolution of the feature maps in layer 5 is increasing back to the same resolution as layer 3. Using information from both layer 3 and 4, the left feature map in layer 5 is progressing the white blurring of the background, while sharpening the edge of the prostate boundary. The middle feature map is highlighting the prostate TZ, while the feature map on the right is showing a combination of a black PZ and a grey TZ as well as  grey background.
\item[layer 6]
This layer progresses the feature maps of layer 5 and sharpens the borders further.
\item[layer 7]
The final layer progresses the features in layer 5 and 6 and sharpens the borders further again. In addition to the two feature maps that segment the background and the prostate TZ (left and middle feature map), feature maps appear that segment the prostate PZ separately (feature map to the right).
\end{itemize}

\begin{figure}[h!]
    \centering
    \includegraphics[width=\textwidth,trim={0cm 0cm 0cm 0cm},clip]{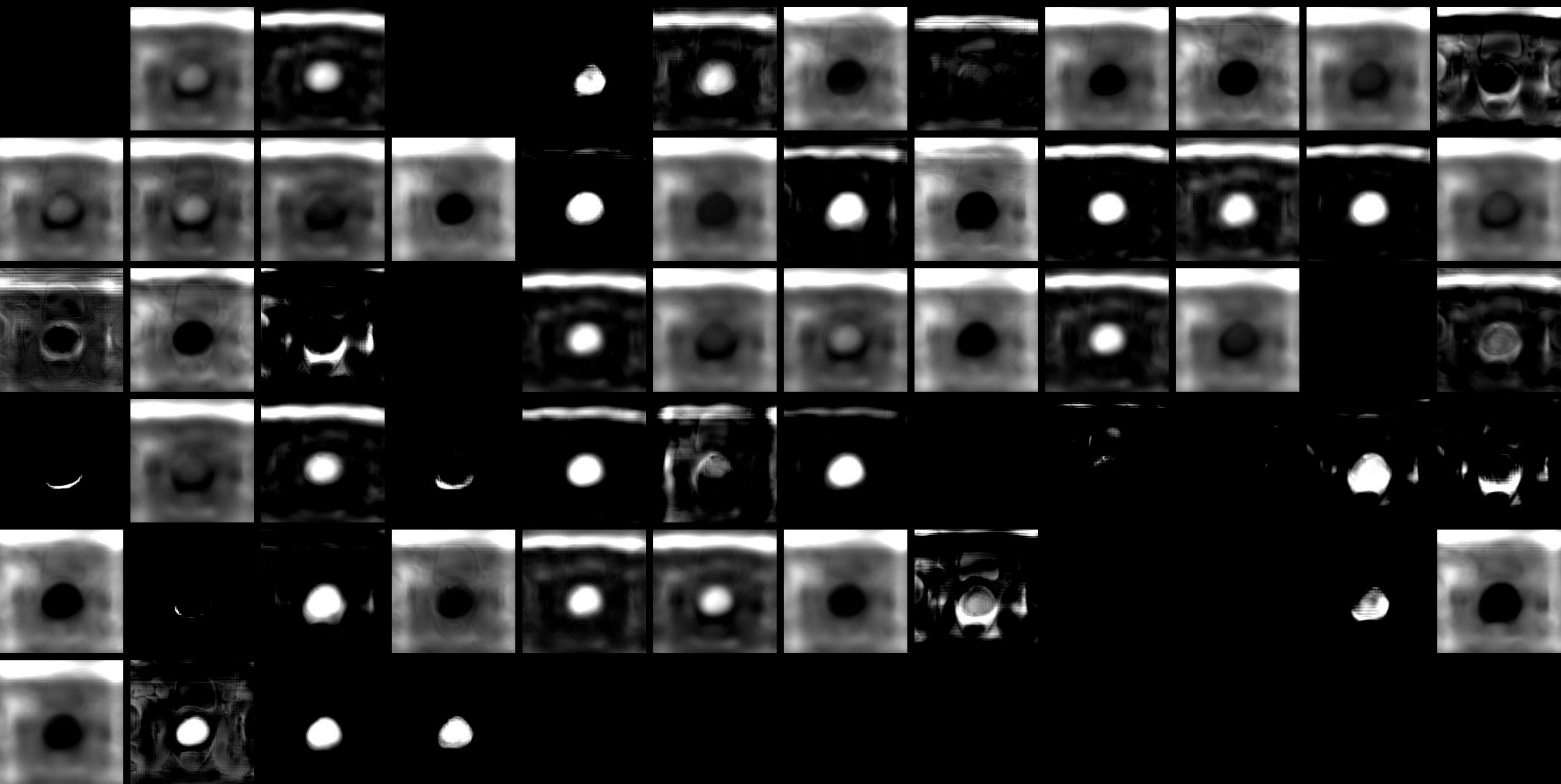}\\
    \vspace{1cm}
    \includegraphics[width=\textwidth,trim={0cm 0cm 0cm 0cm},clip]{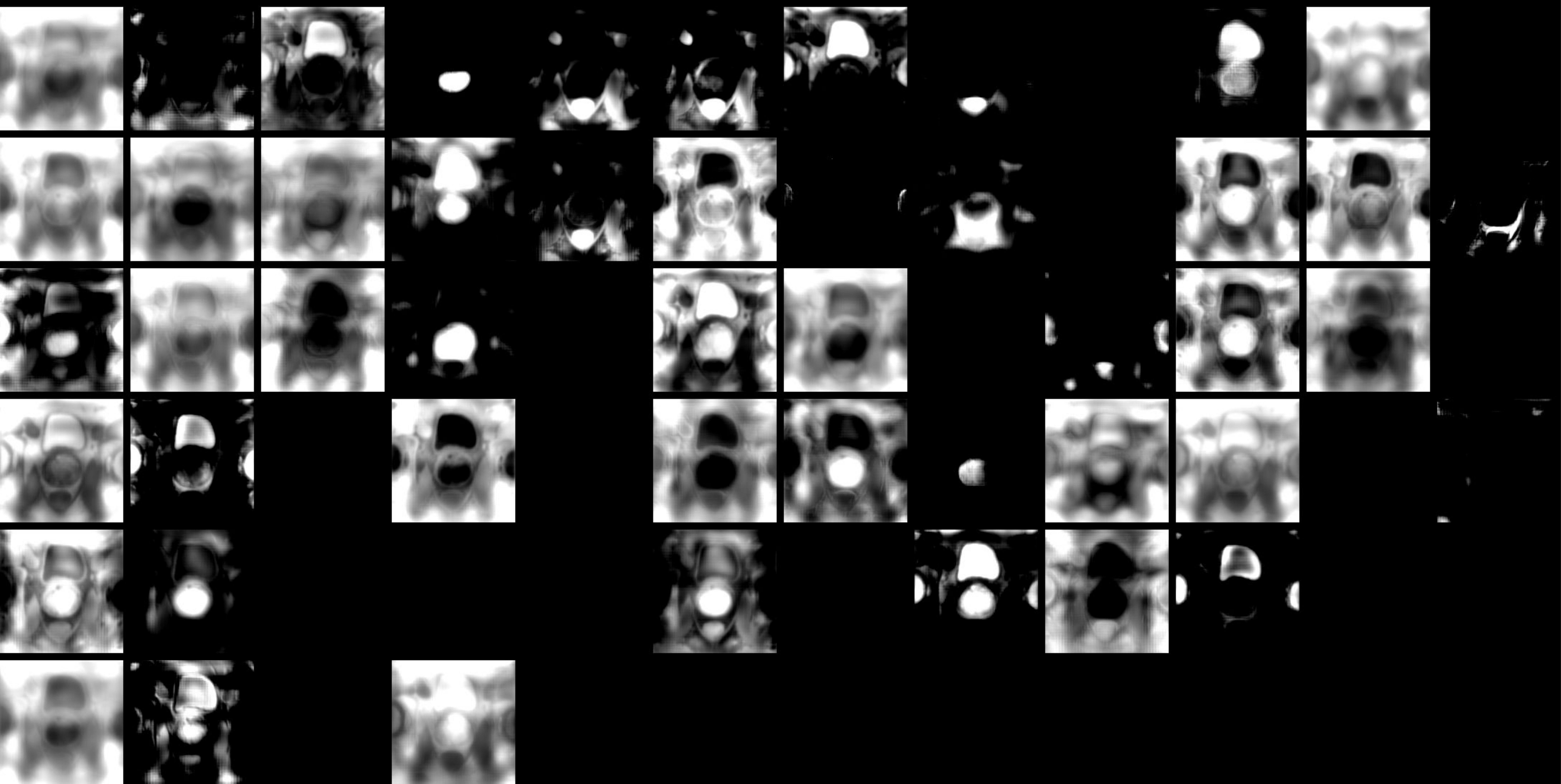}
\caption{Map of all 64 features for the layer conv7b, for the case where 2 labels (top) or 6 labels (bottom) are segmented by the aniso-3DUnet model.}
\label{fig:feature_map_conv7b_2_lbl_5lbl}
\end{figure}

The final, seventh, layer is preparing well for the segmentation into prostate TZ, PZ, and background in the softmax layer. To find out why the performance of the model in segmenting PZ is better for 2-label segmentation than for 6-label segmentation, we plot in Figure \ref{fig:feature_map_conv7b_2_lbl_5lbl} all the feature maps for layer 7 for both cases. The difference observed is that for the 6-label segmentation case, feature maps that were coding for a separate prostate PZ in the 2-label case are replaced by feature maps that are coding for combinations of the other organs (bladder, rectum, and femur bones).

\section{Conclusions and discussion}
We have evaluated a 3D U-net neural network \cite{Cicek} for automatic segmentation of both prostate TZ and PZ in MRI scans and have found that it could achieve segmentations with dice scores of 0.85 for TZ and 0.60 for PZ. This is a few percent higher than a recently published ATLAS method \cite{Padgett2016} (0.83 for prostate, 0.57 for PZ) and a 2-stage 2D U-net method \cite{Clark2017} (0.82 for prostate and 0.77 for TZ).

We explored two ideas for improving the 3D U-net performance, that make use of characteristics specific to MRI images. One characteristic is the anisotropy of the MRI volumes for which we tested two architectures: an anisotropic network architecture, aniso-3DUNET, that reflects the anisotropy in the MRI volumes (see Figure \ref{fig:anisotropic3Dunet}) and a more isotropic architecture (see Figure \ref{fig:isotropic3Dunet}), iso-3DUNET. The aniso-3DUNET performs slightly better (0.60 versus 0.57), but we consider the difference marginal because it is of the same order of magnitude as the standard error in the average dice scores of for all volumes.

Another characteristic is that the images always have a fixed topology, that is given by the anatomy of the organ of interest and surrounding tissues. We have tested whether training the network to segment additional tissues (bladder, rectum and femur bones) would improve the segmentation, but against our expectations this significantly decreases the dice score by 0.07.

An explanation for the decreased segmentation performance of the network when segmenting more tissues was found by plotting the activations of the feature maps for each of the 3D U-net layers for one of the test MRI volumes. This clearly showed that when only the prostate PZ and TZ are segmented, the last layer of the network learns features dedicated to each label: background, TZ and PZ. When more tissues are segmented, the features dedicated to PZ are replaced by combination features of other tissues. An increase in the number of filters might allow for dedicated feature maps, but at the cost of significant increase of GPU memory requirements. And the added value compared to the 2-label case may be limited, considering that Figure \ref{fig:feature_maps_2lbl} shows that the network is already taking surrounding shapes into account in earlier layers without explicitly segmenting them in the final layer.

Visualizing the feature map activations also provides insights in how the 3D U-net manages to segment the images. In the first 3 layers different shapes are detected utilizing distinctive local features. These local features are edges and intensities, but also thick versus thin shapes. The coarsest fourth layer combines feature maps that have the area of interest in common and that blur out the background when superimposed. The prostate PZ is too thin to be coded in the fourth layer, and the distinction between TZ and PZ is detected in the third layer. In the last layers 5, 6 and 7, these areas of interest are sharpened by overlaying the coarse feature maps with higher resolution feature maps from earlier layers that detect the edges of the segmented zones. Finally feature maps are combined to form feature maps that separately detect each label in the same resolution as the original image.

\newpage

\printbibliography

\newpage

\setcounter{section}{0}
\renewcommand\thesection{\Alph{section}}

\section{Appendices}

\subsection{3D U-net feature maps for 2-label case}

\begin{figure}[h!]
   \centering
    \includegraphics[width=\textwidth,trim={0cm 0cm 0cm 0cm},clip]{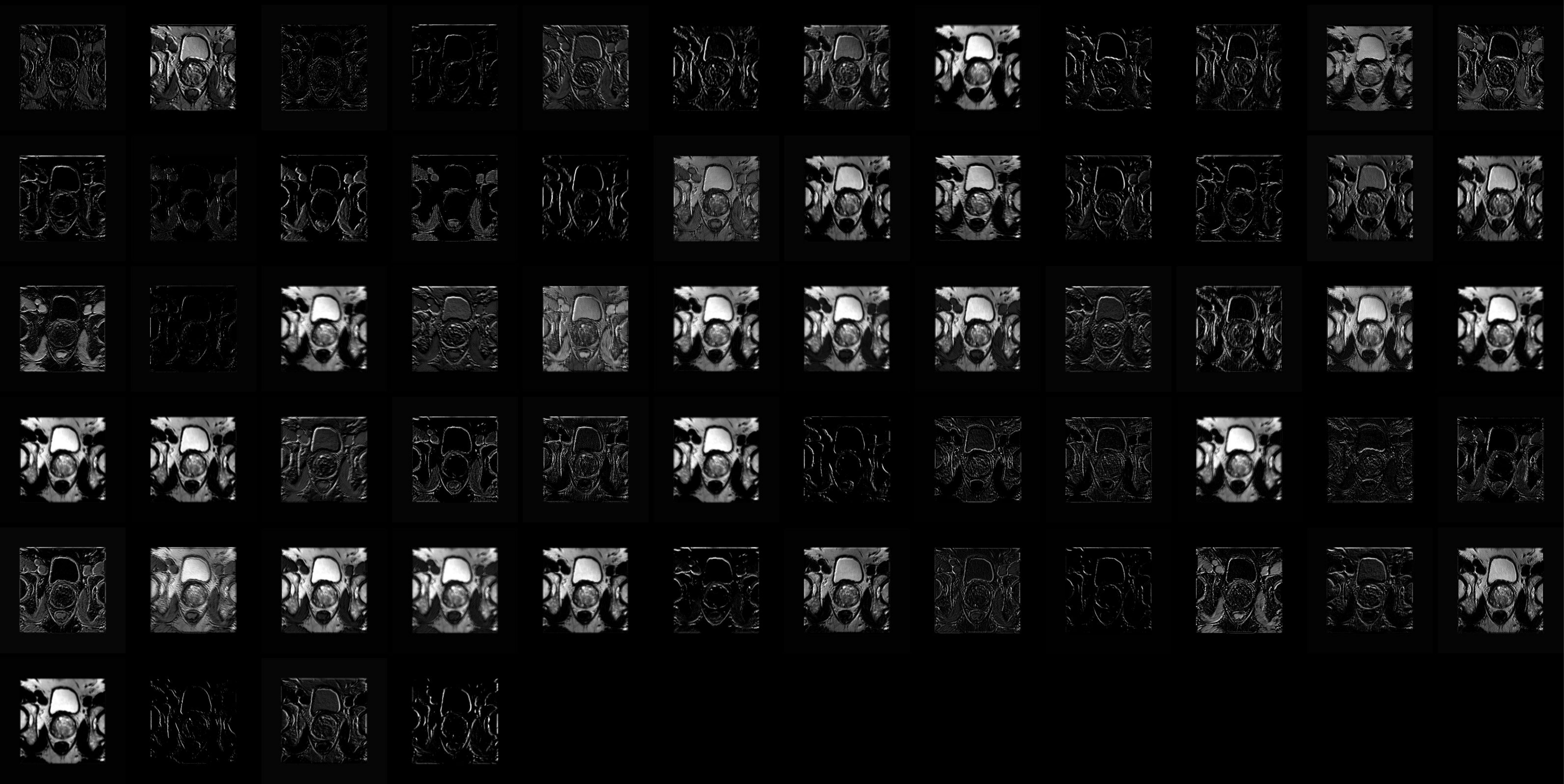}
    \caption{64 features of layer 1}
\end{figure}
\begin{figure}[h!]
   \centering
    \includegraphics[width=\textwidth,trim={0cm 0cm 0cm 0cm},clip]{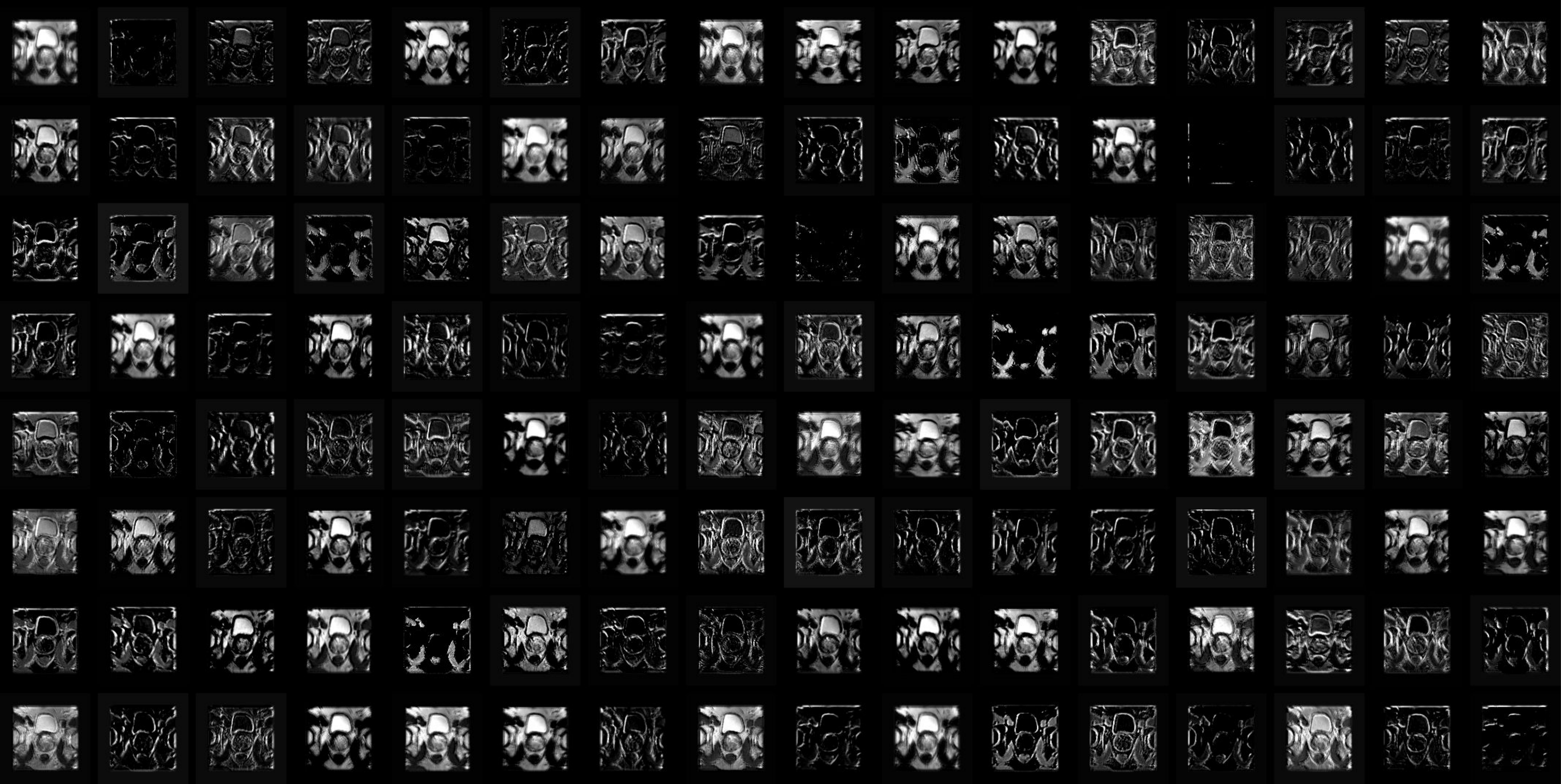}
    \caption{128 features of layer 2}
\end{figure}
\begin{figure}[h!]
   \centering
    \includegraphics[width=\textwidth,trim={0cm 0cm 0cm 0cm},clip]{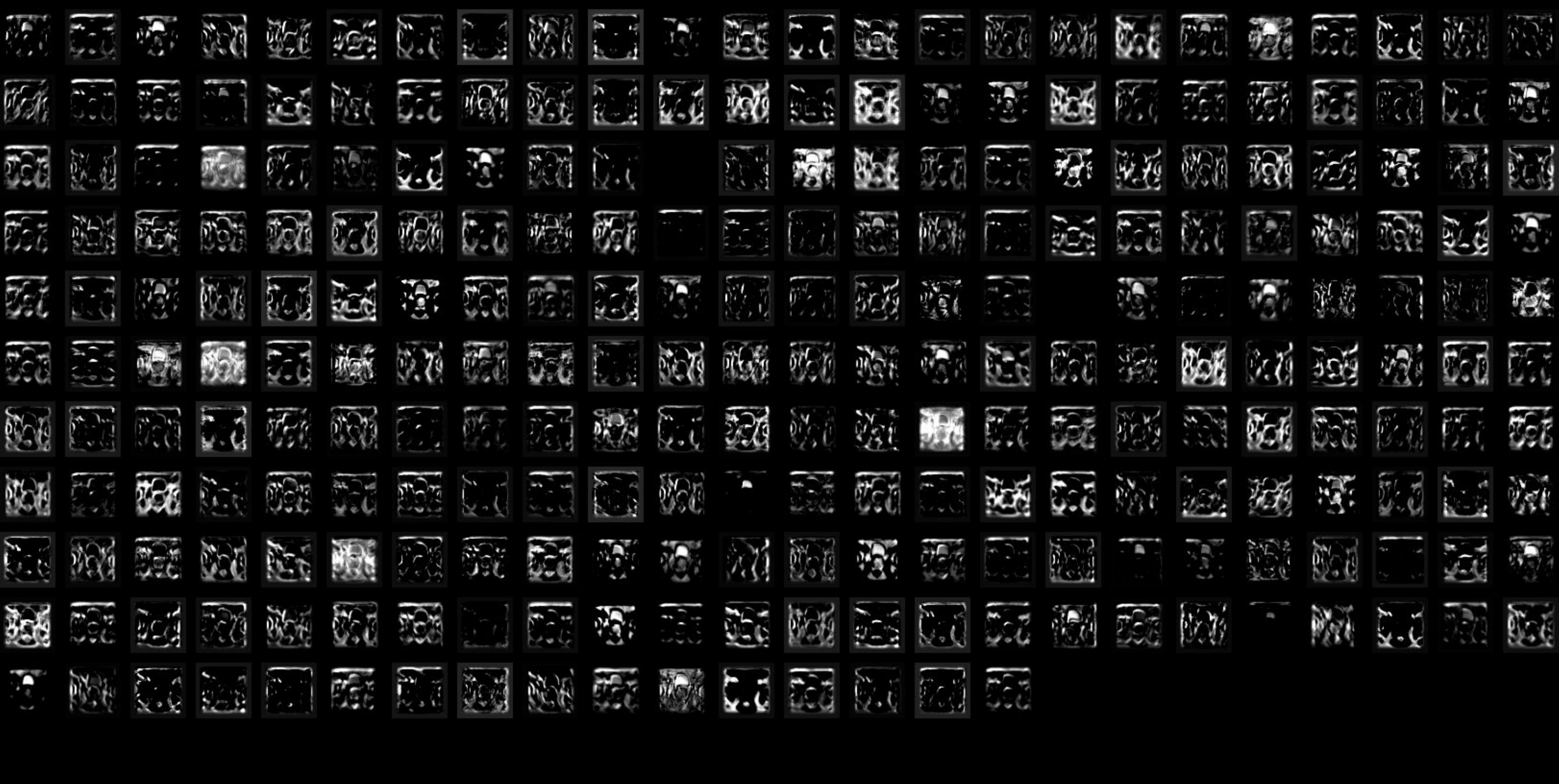}
    \caption{256 features of layer 3}
\end{figure}
\begin{figure}[h!]
   \centering
    \includegraphics[width=\textwidth,trim={0cm 0cm 0cm 0cm},clip]{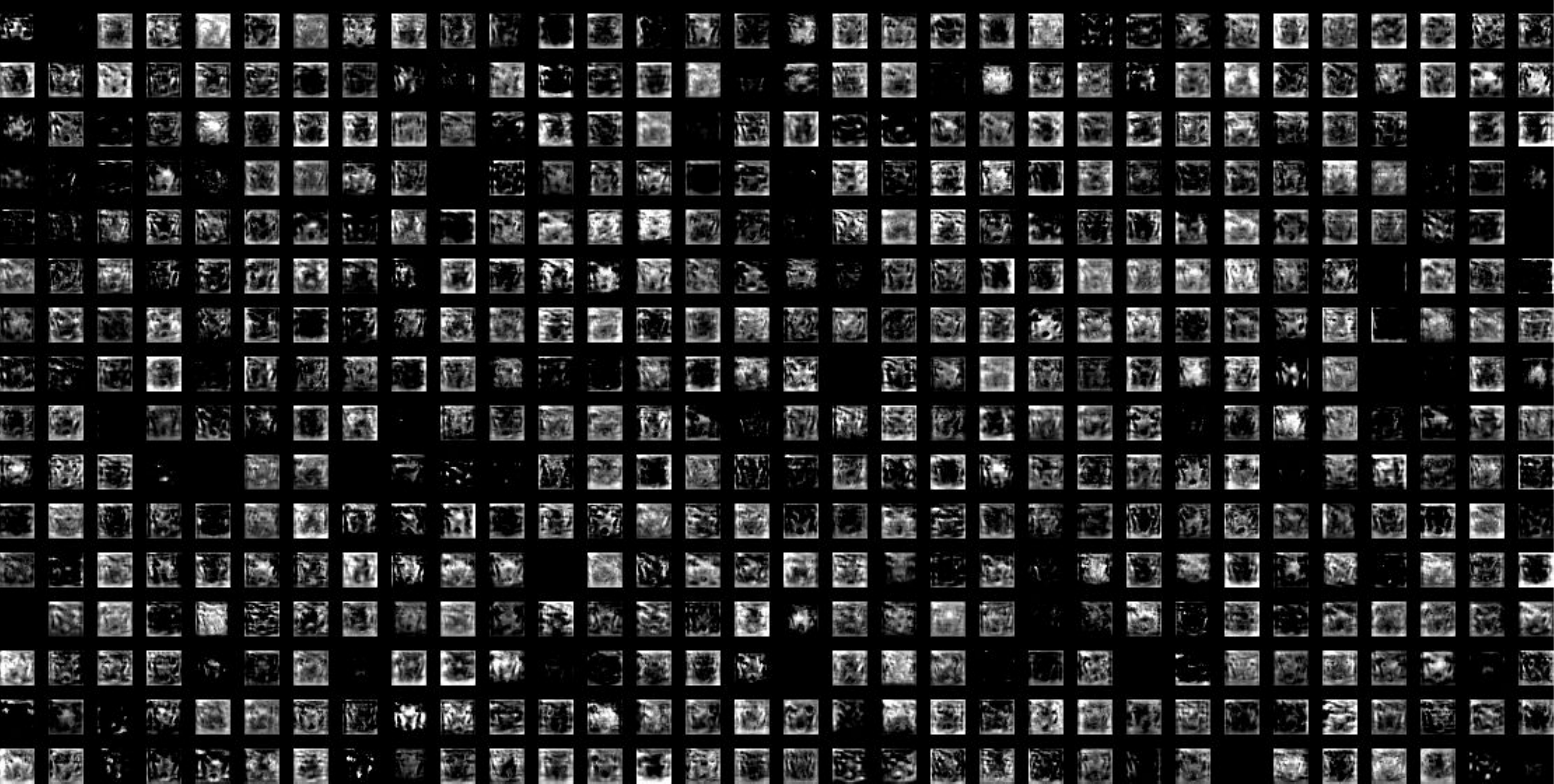}
    \caption{512 features of layer 4}
    \end{figure}
\begin{figure}[h!]
   \centering
    \includegraphics[width=\textwidth,trim={0cm 0cm 0cm 0cm},clip]{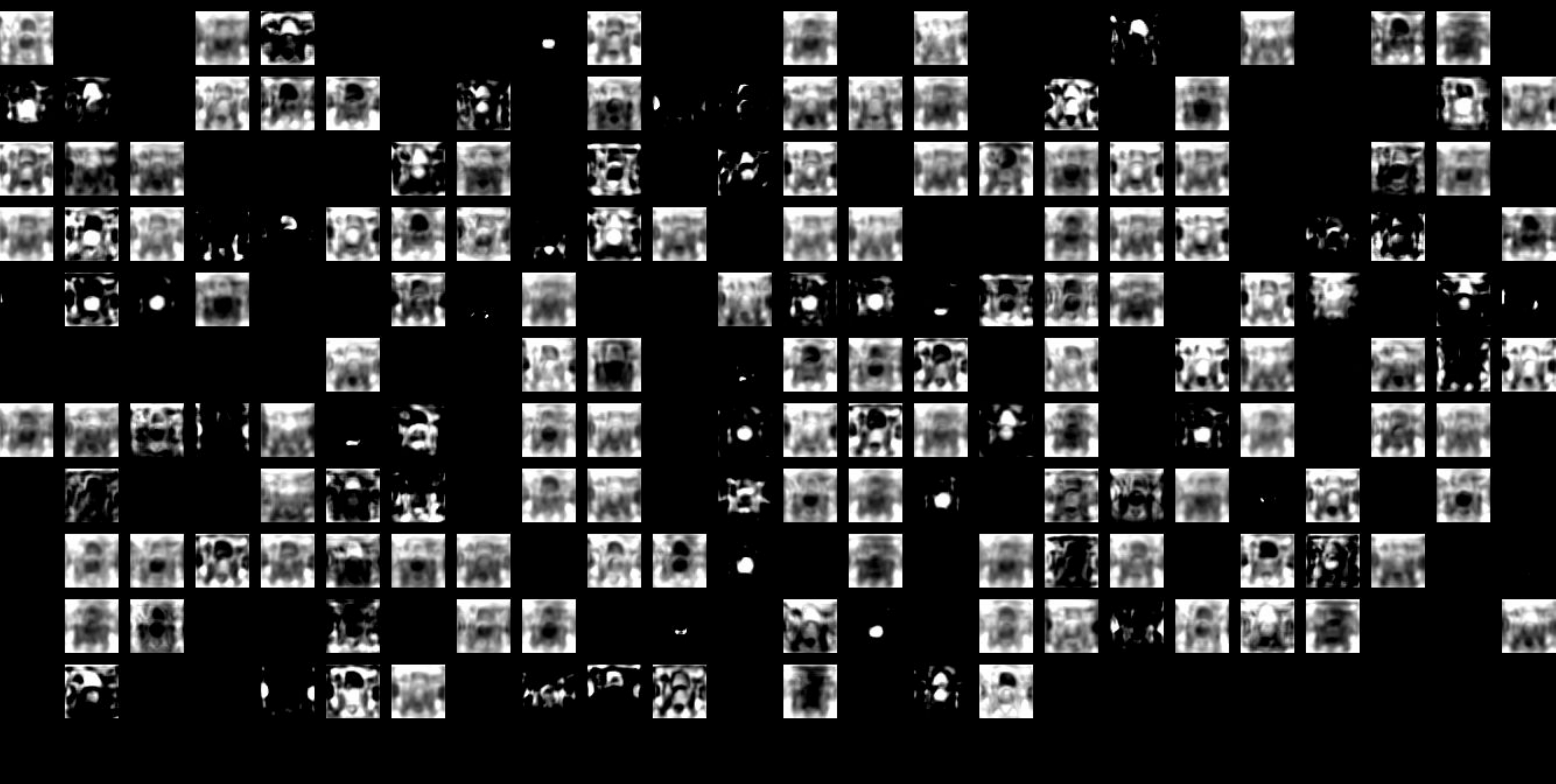}
    \caption{256 features of layer 5}
    \end{figure}
\begin{figure}[h!]
   \centering
    \includegraphics[width=\textwidth,trim={0cm 0cm 0cm 0cm},clip]{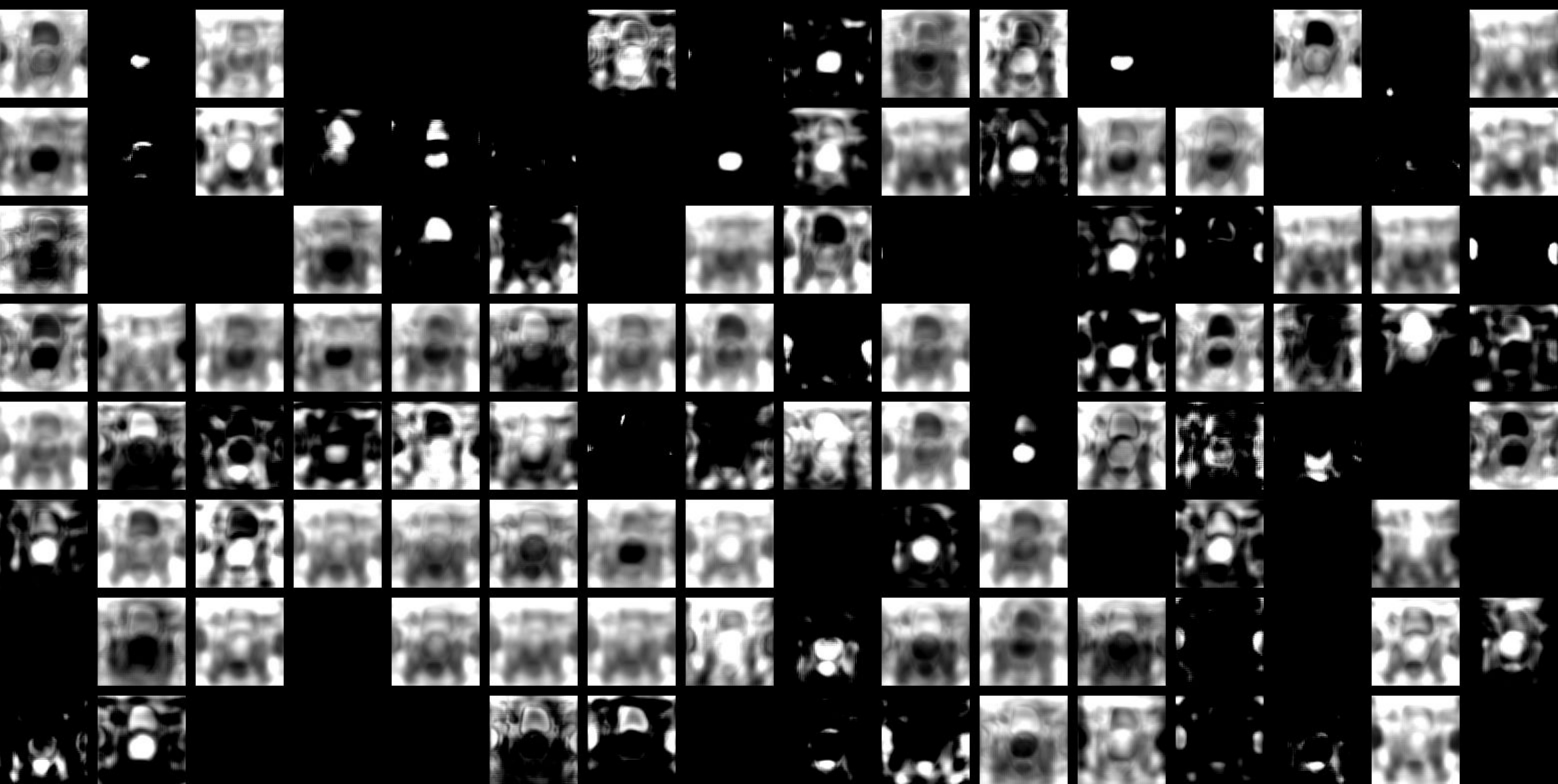}
    \caption{128 features of layer 6}
    \end{figure}
\begin{figure}[h!]
   \centering
    \includegraphics[width=\textwidth,trim={0cm 0cm 0cm 0cm},clip]{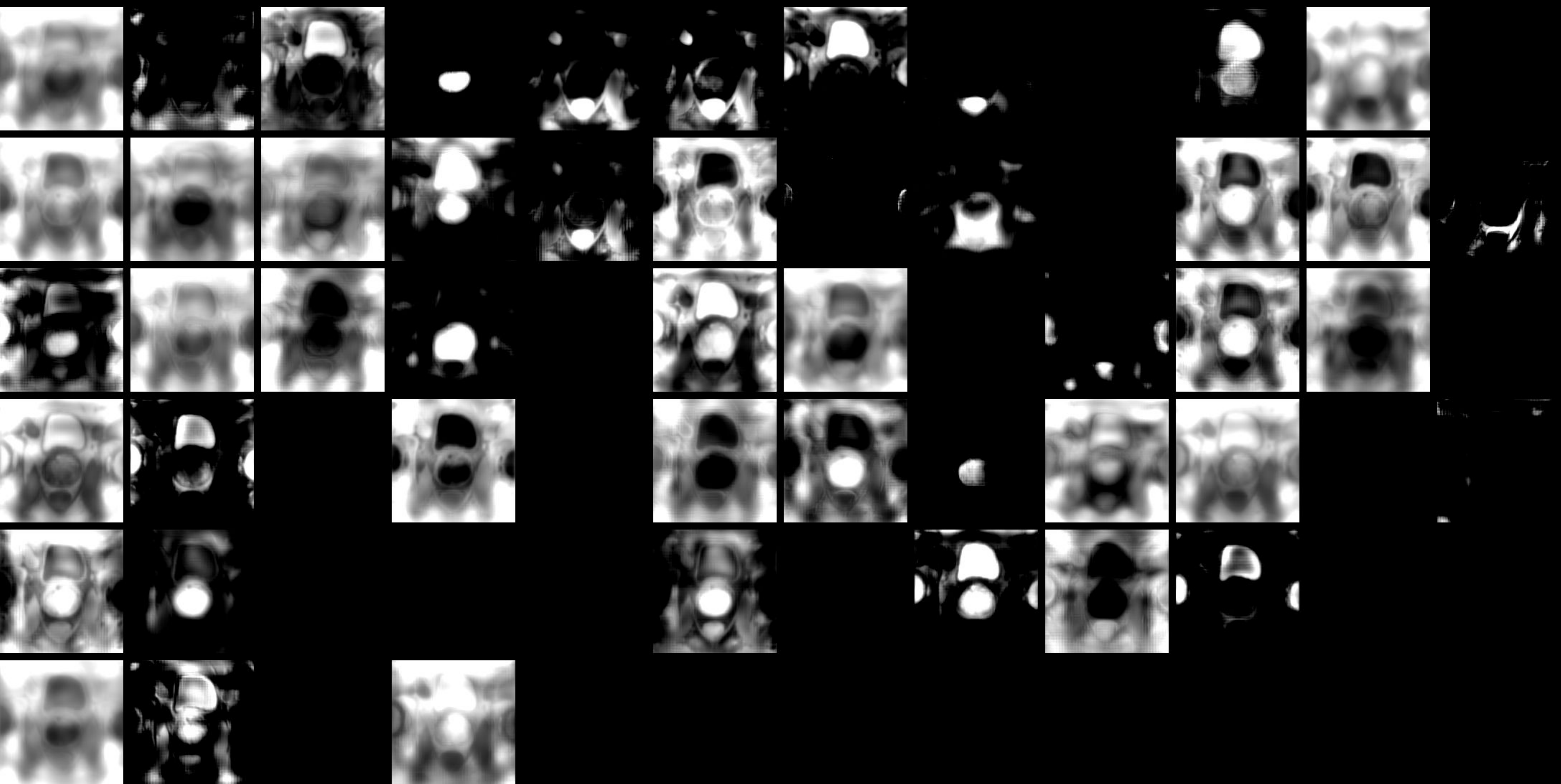}
    \caption{64 features of layer 7}
    \end{figure}

\newpage

\subsection{3D U-net feature maps for 6-label case}

\begin{figure}[h!]
   \centering
    \includegraphics[width=\textwidth,trim={0cm 0cm 0cm 0cm},clip]{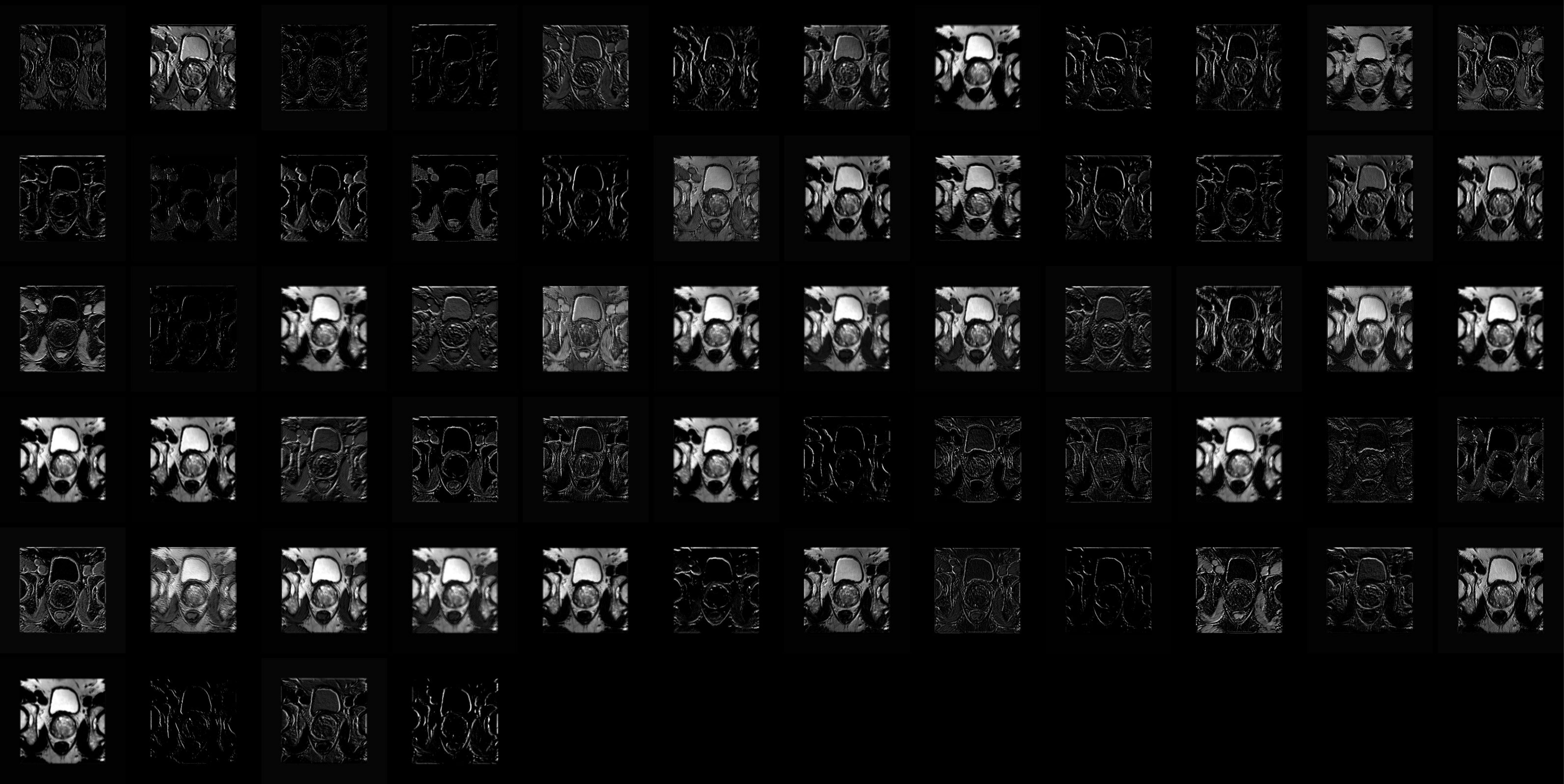}
    \caption{64 features of layer 1}
\end{figure}
\begin{figure}[h!]
   \centering
    \includegraphics[width=\textwidth,trim={0cm 0cm 0cm 0cm},clip]{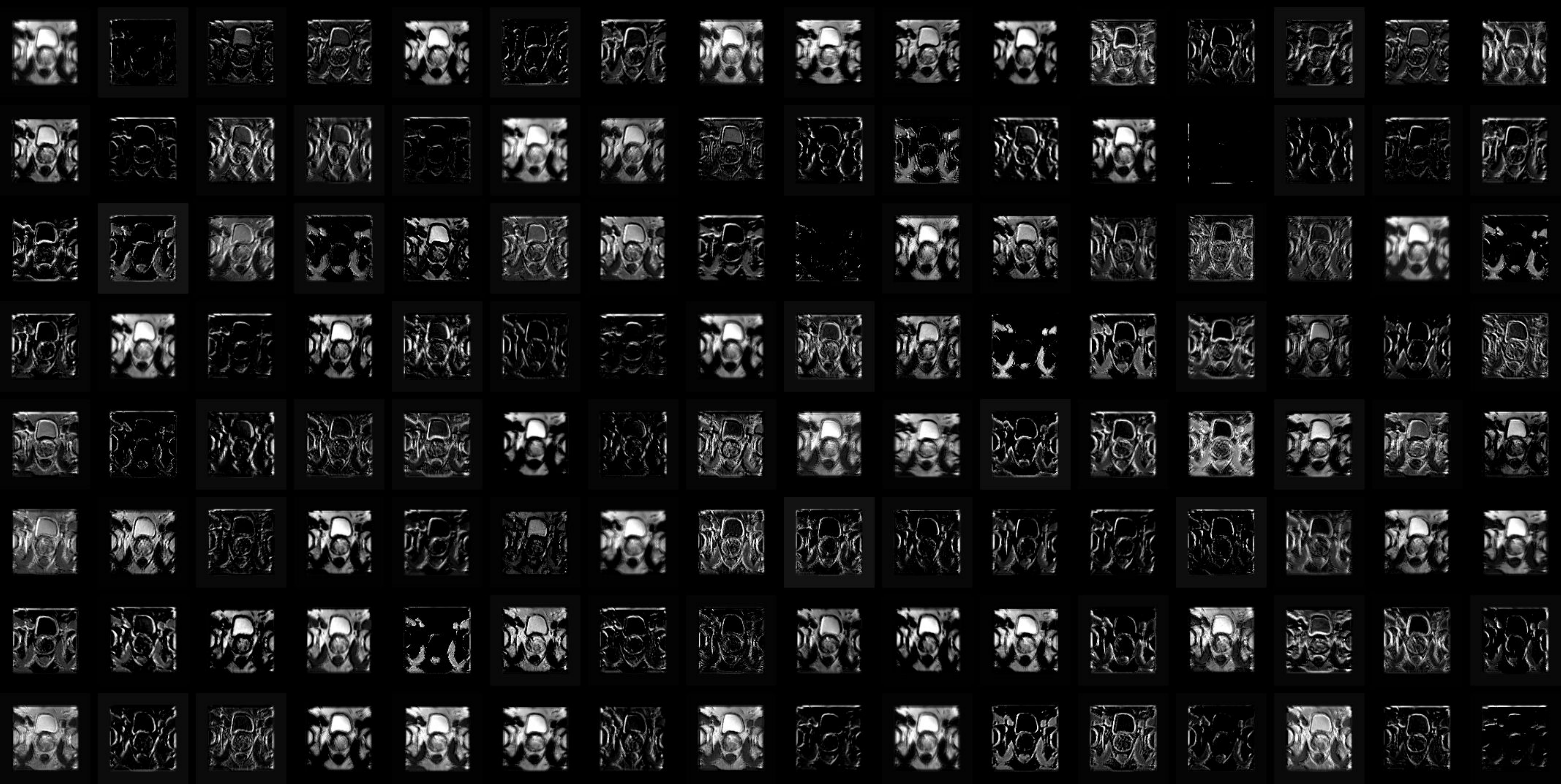}
    \caption{128 features of layer 2}
\end{figure}
\begin{figure}[h!]
   \centering
    \includegraphics[width=\textwidth,trim={0cm 0cm 0cm 0cm},clip]{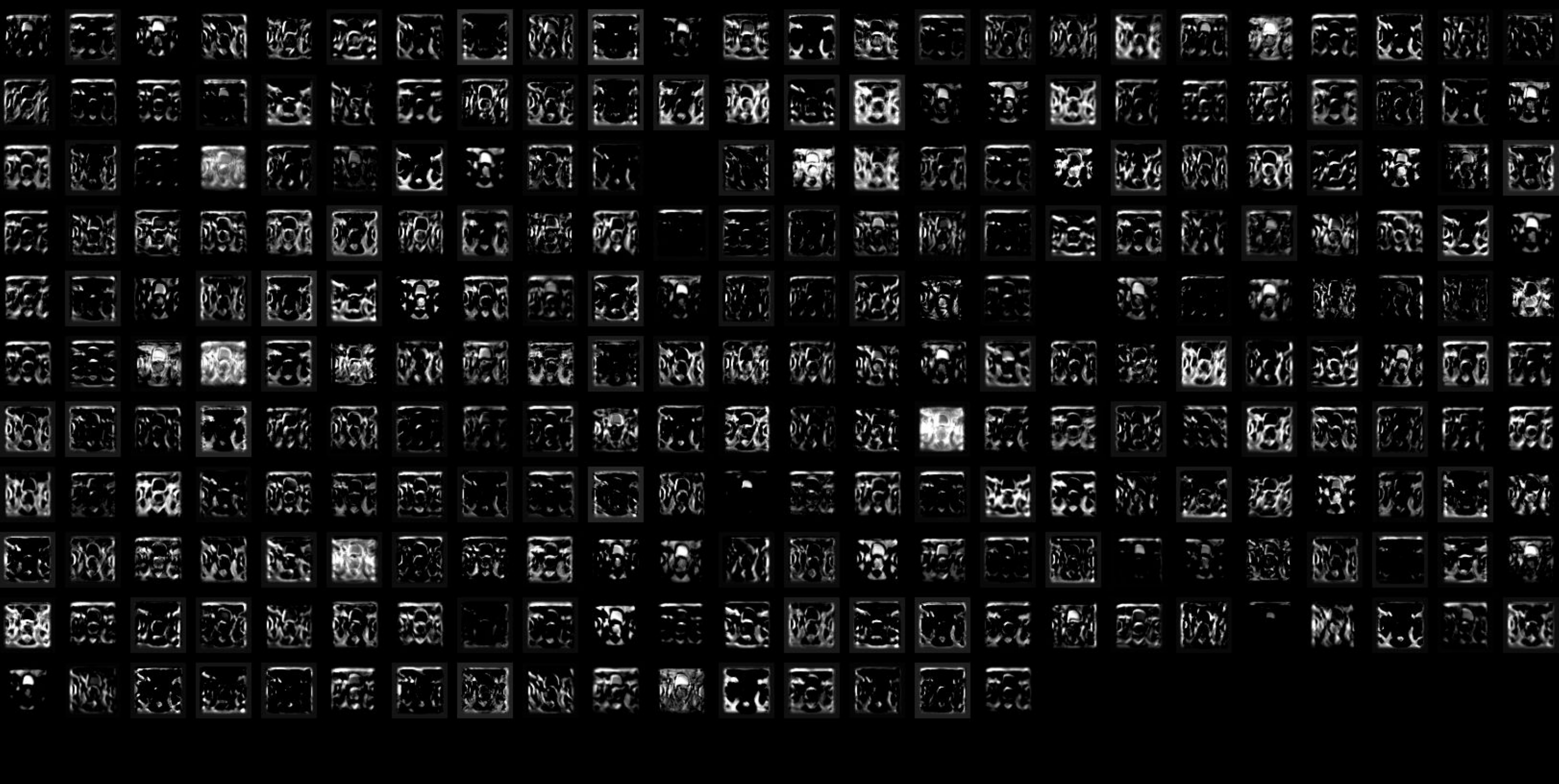}
    \caption{256 features of layer 3}
\end{figure}
\begin{figure}[h!]
   \centering
    \includegraphics[width=\textwidth,trim={0cm 0cm 0cm 0cm},clip]{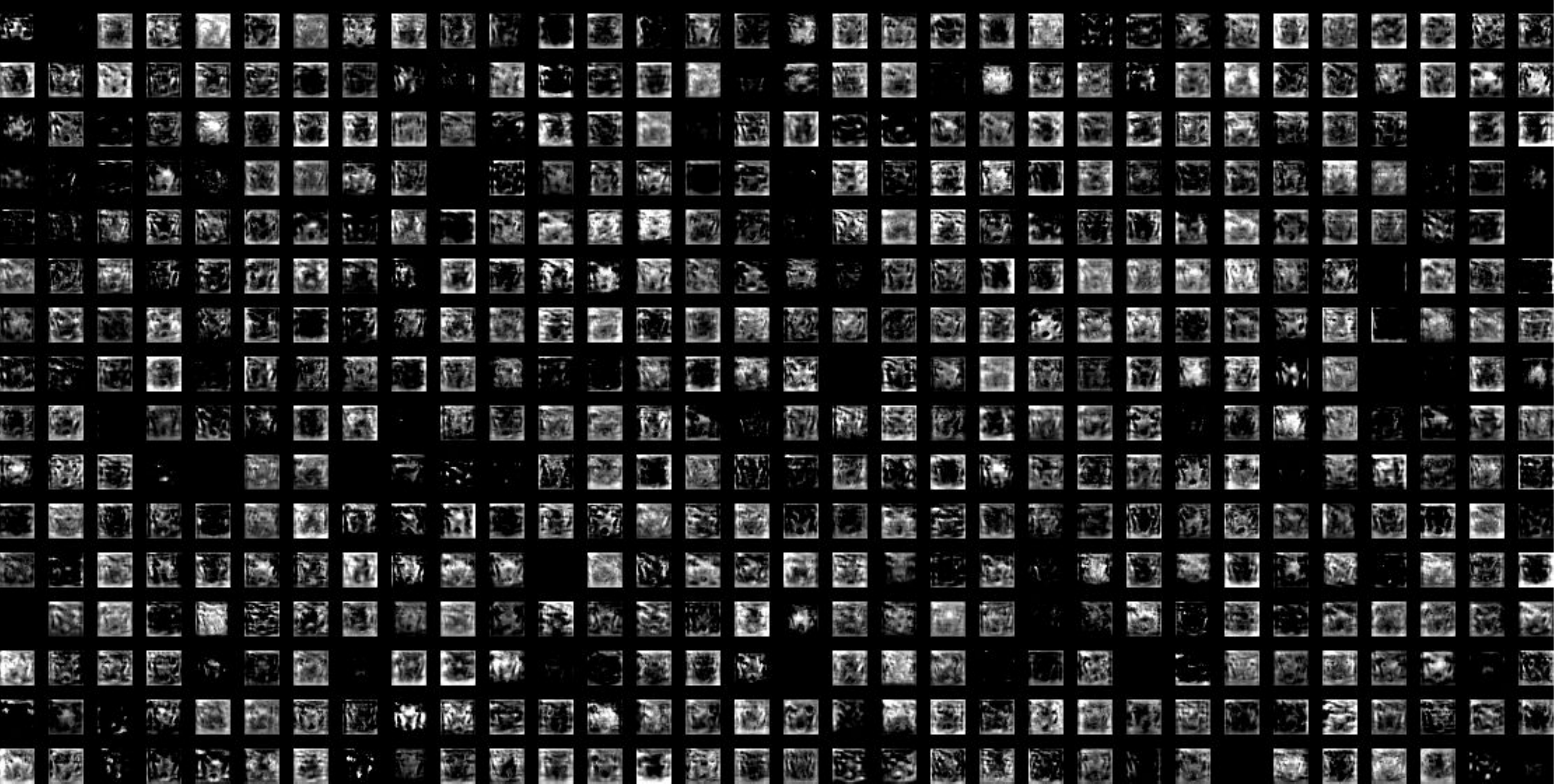}
    \caption{512 features of layer 4}
    \end{figure}
\begin{figure}[h!]
   \centering
    \includegraphics[width=\textwidth,trim={0cm 0cm 0cm 0cm},clip]{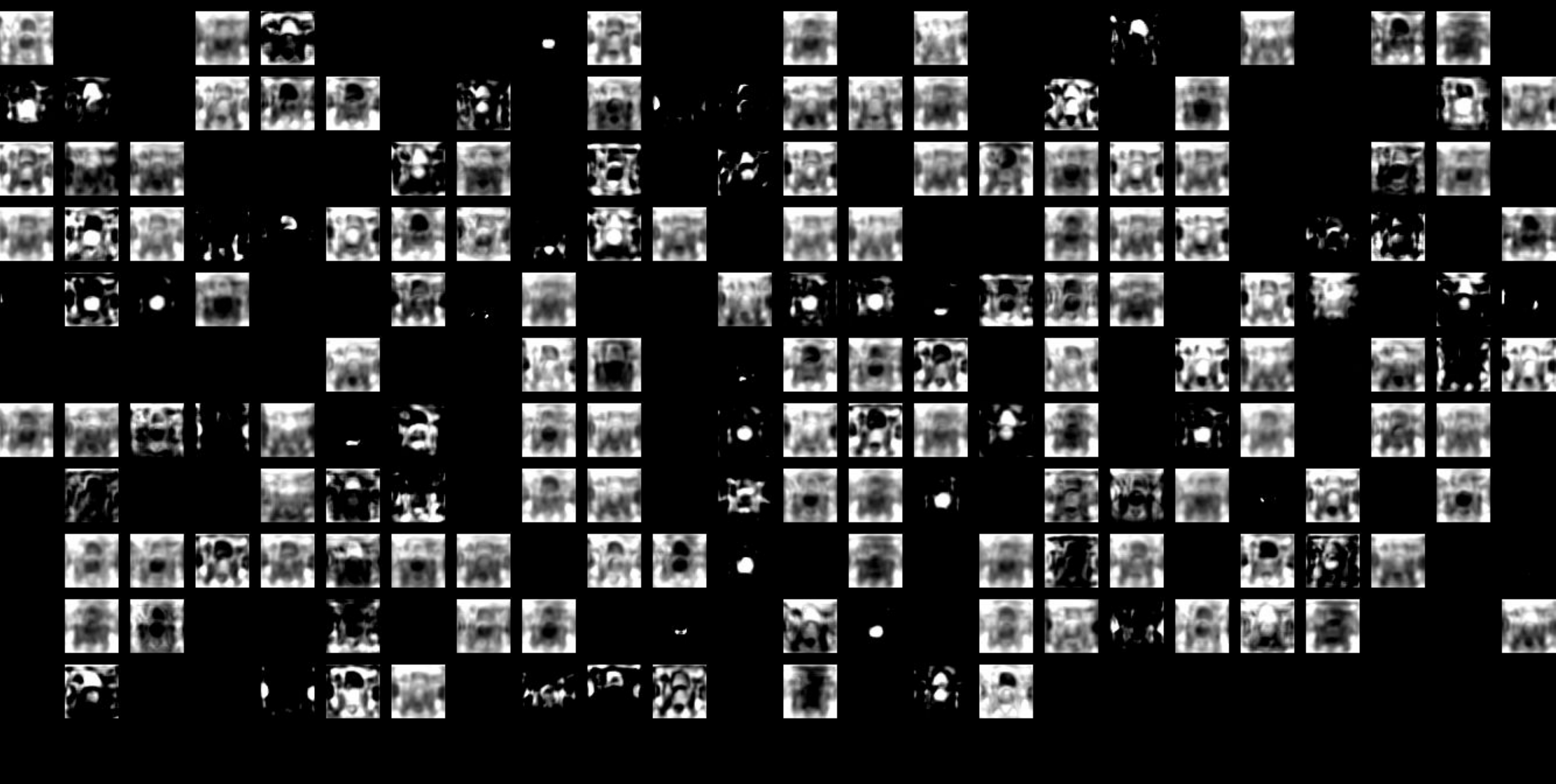}
    \caption{256 features of layer 5}
    \end{figure}
\begin{figure}[h!]
   \centering
    \includegraphics[width=\textwidth,trim={0cm 0cm 0cm 0cm},clip]{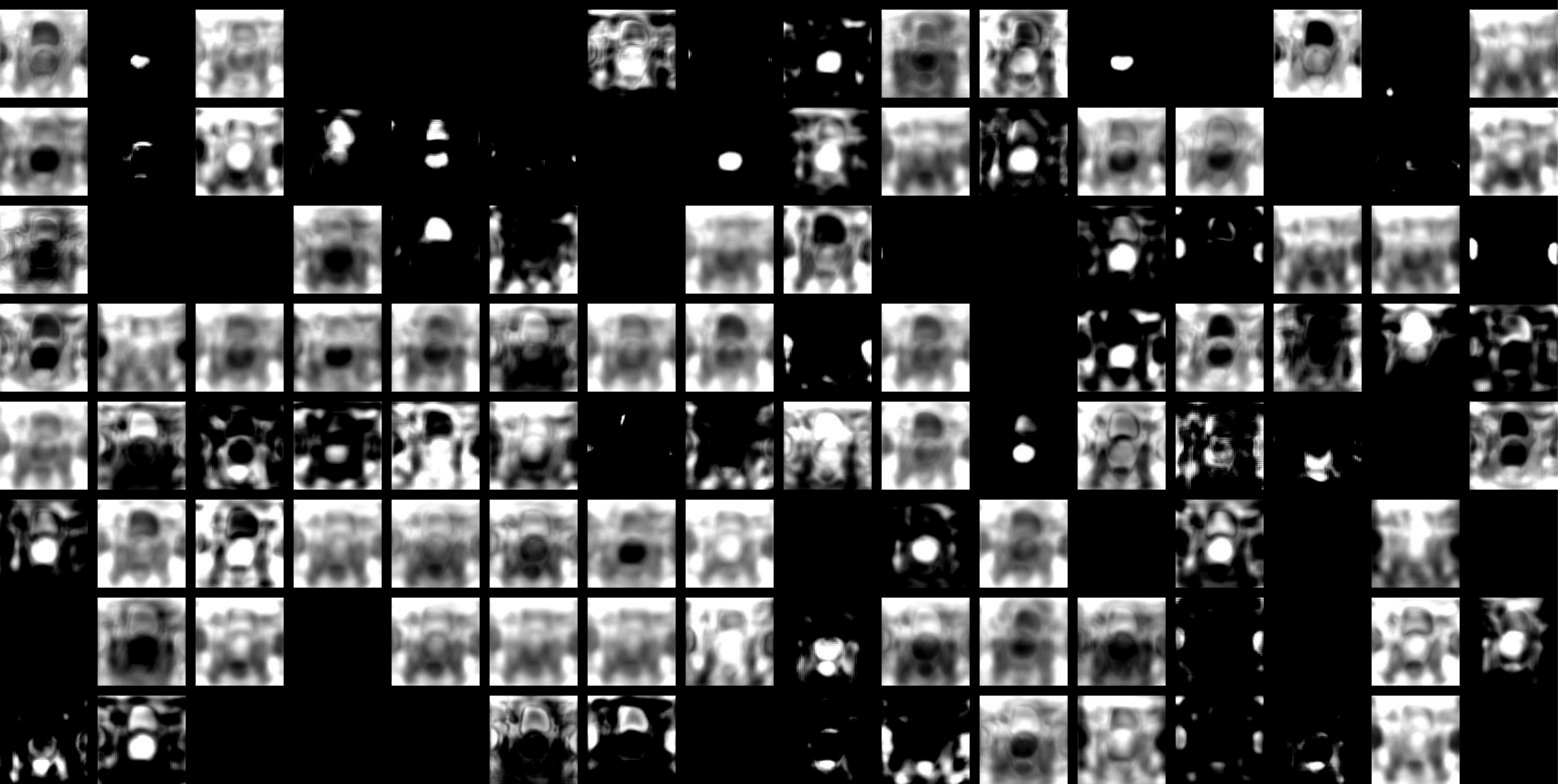}
    \caption{128 features of layer 6}
    \end{figure}
\begin{figure}[h!]
   \centering
    \includegraphics[width=\textwidth,trim={0cm 0cm 0cm 0cm},clip]{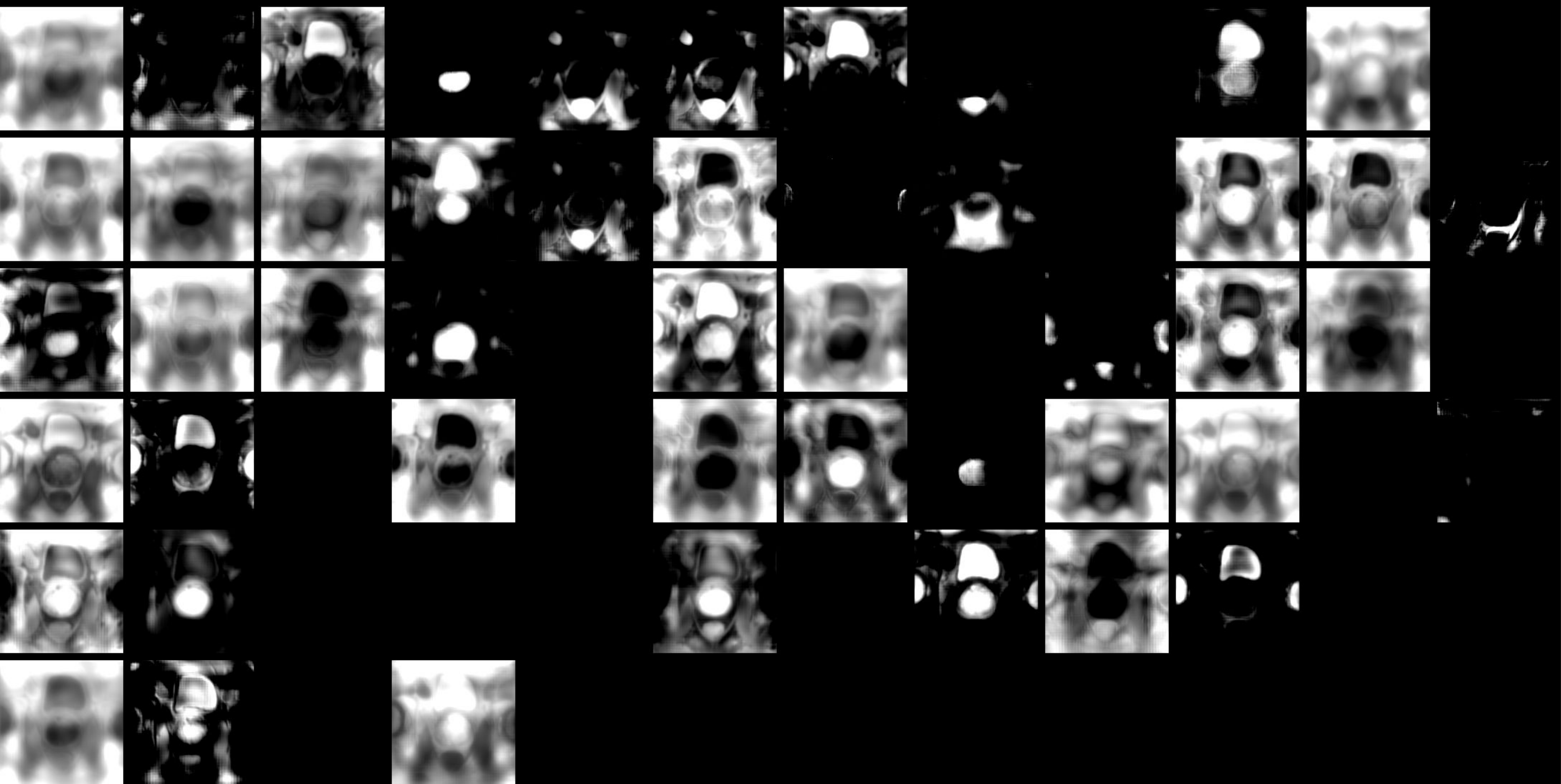}
    \caption{64 features of layer 7}
    \end{figure}

\end{document}